\documentclass[twocolumn,10pt]{article}
\usepackage[letterpaper,margin=0.75in]{geometry}
\usepackage{mathptmx}                 
\usepackage{titlesec}
\titleformat*{\section}{\large\bfseries\raggedright}
\titleformat*{\subsection}{\normalsize\bfseries\raggedright}
\titleformat*{\subsubsection}{\normalsize\bfseries\raggedright}
\titlespacing*{\section}{0pt}{2.0ex plus .4ex minus .2ex}{1.0ex plus .2ex}
\titlespacing*{\subsection}{0pt}{1.5ex plus .3ex minus .2ex}{0.7ex plus .1ex}
\titlespacing*{\subsubsection}{0pt}{1.2ex plus .3ex minus .2ex}{0.5ex plus .1ex}
\renewcommand{\footnoterule}{\kern-3pt\hbox{}\kern2.6pt}

\usepackage[T1]{fontenc}
\usepackage[utf8]{inputenc}
\usepackage{textcomp}
\usepackage{amsmath,amssymb,amsfonts}
\usepackage{algorithm}      
\usepackage{algorithmic}
\IfFileExists{placeins.sty}{\usepackage{placeins}}{\providecommand{\FloatBarrier}{}}
\usepackage{cite}

\usepackage{graphicx}
\usepackage{booktabs}
\usepackage{array}
\usepackage{xcolor}
\usepackage{tikz}
\usetikzlibrary{shapes.geometric,arrows.meta,positioning,fit,backgrounds,calc}
\IfFileExists{microtype.sty}{\usepackage{microtype}}{}
\usepackage{hyperref}
\definecolor{citeblue}{HTML}{0645AD}
\hypersetup{colorlinks=true, linkcolor=citeblue, citecolor=citeblue, urlcolor=black}

\def\BibTeX{{\rm B\kern-.05em{\sc i\kern-.025em b}\kern-.08em
    T\kern-.1667em\lower.7ex\hbox{E}\kern-.125emX}}

\definecolor{ink}{HTML}{171717}
\definecolor{muted}{HTML}{666666}
\definecolor{oblue}{HTML}{0070F3}
\definecolor{ogreen}{HTML}{00A862}
\definecolor{oamber}{HTML}{F5A623}
\definecolor{ored}{HTML}{E5484D}
\definecolor{oviolet}{HTML}{7928CA}

\tikzset{
  stage/.style={rectangle, rounded corners=2pt, draw=ink, line width=0.6pt,
    fill=white, align=center, font=\scriptsize, minimum height=7mm, inner sep=3pt},
  data/.style={rectangle, draw=oblue, line width=0.6pt, fill=oblue!6,
    align=center, font=\scriptsize, minimum height=7mm, inner sep=3pt},
  agent/.style={rectangle, rounded corners=3pt, draw=ogreen, line width=0.7pt,
    fill=ogreen!6, align=center, font=\scriptsize, minimum height=8mm, inner sep=3pt},
  lever/.style={rectangle, rounded corners=2pt, draw=oviolet, line width=0.6pt,
    fill=oviolet!7, align=center, font=\scriptsize\itshape, inner sep=3pt},
  flow/.style={-{Stealth[length=2mm]}, line width=0.6pt, draw=muted},
}

\newcommand{\CONFTWOHUNDRED}{$0.990$}
\newcommand{\CONFTWOHUNDREDSHORT}{0.990}
\newcommand{\ABLpanelZero}{$0.500$}
\newcommand{\ABLpanelFS}{$0.983$}
\newcommand{\ABLpanelSC}{$0.483$}
\newcommand{\ABLpanelBoth}{$\mathbf{1.000}$}
\newcommand{\ABLsingleZero}{$0.400$}
\newcommand{\ABLsingleBoth}{$0.983$}
\newcommand{\GEMzeroDet}{1.000}
\newcommand{\GEMzeroCls}{0.715}
\newcommand{\GEMenhCls}{0.995}    
\newcommand{\SWEEPcells}{9}          
\newcommand{\SWEEPfaults}{3}
\newcommand{\SWEEPseeds}{3}
\newcommand{\SWEEPpolls}{41}         
\newcommand{\SWEEPdetRate}{1.00}     
\newcommand{\SWEEPlatMed}{29.5}      
\newcommand{\SWEEPlatMax}{59.5}      
\newcommand{\SWEEPfloorN}{8}         
\newcommand{\SWEEPfa}{0}             
\newcommand{\SWEEPsingleFA}{11}      

\newcommand{\ENGmulti}{0.915}        
\newcommand{\ENGmacro}{0.898}        
\newcommand{\ENGthreshMulti}{0.245}  
\newcommand{\ENGnoiseEng}{0.101}     
\newcommand{\ENGnoiseRF}{0.701}      

\begin{document}

\title{\huge\bfseries Onnes: A Physics-Grounded Multi-Agent LLM Simulator
for Cryogenic Fault Diagnosis in Quantum Computing Infrastructure}

\author{Praneeth Narisetty, Uday Kumar Reddy Kattamanchi, Shiva Nagendra Babu Kore\\[3pt]
\href{https://onnes.ai}{\textbf{Onnes Research}} \textbar{} San Francisco, CA\\[2pt]
\{praneeth, uday, shiva, research\}@onnes.ai}
\date{}

\maketitle
\vspace*{-3.3em}   

\begin{abstract}
Dilution refrigerators are the enabling infrastructure of superconducting quantum
computers, yet their fault diagnosis is still dominated by threshold alarms that
report \emph{that} something is wrong, not \emph{what}. We present \textbf{Onnes}, and
our headline result is parity without training: curated contrastive few-shot
demonstrations and self-consistency voting raise a zero-shot LLM agent panel's cryogenic
fault-classification accuracy from $0.685$ to $0.990$ --- matching a supervised classifier
($0.985$) with no parameter updates and only six labeled demonstrations --- while a
detector trained purely on real BlueFors telemetry posts a genuine real-hardware
false-alarm rate of $6.4\%$. Onnes is a
physics-grounded digital-twin \emph{simulator} of a dilution refrigerator (a
forward physics model with a learned real-fridge noise fingerprint, not a
hardware-coupled bidirectional twin) that drives a live multi-agent
large-language-model (LLM) operations layer. With it we run one of the first
controlled head-to-head comparisons between a zero-shot LLM agent panel and a
supervised machine-learning (ML) classifier on honestdiagnosis. The twin
couples a real dilution-cooling floor, a noise-and-correlation fingerprint learned
from real BlueFors logs, and six physics-grounded fault classes, three of them
engineered to overlap on temperature but separate on flow and pressure. Across a
1000-turn evaluation, the zero-shot panel shows no statistically significant difference
from the classifier on fault detection but trails on classification, its errors
concentrating on the engineered confusable
faults; the in-context lift closes exactly those cells, and an ablation attributes the gain
almost entirely to the demonstrations. Run as a continuous monitor across a
nine-run fault$\times$seed sweep, the agent catches every developing fault within one
poll interval, and a confidence gate suppresses pre-onset false alarms whose rate we find
to be backend-dependent. As a first sim-to-real check we take the initial stage of our
validation plan off the roadmap and run it: the same detector reaches
$100\%$ recall on physics faults injected onto real held-out windows, so its low false-alarm
rate does not come at the cost of missed faults. All numbers are drawn verbatim
from released run logs.
\end{abstract}

\vspace{0.5\baselineskip}

\section{Introduction}
Superconducting and spin-qubit quantum processors operate at the base temperature
of a dilution refrigerator, typically $10$--$35$\,mK at the mixing chamber (MXC).
The operational reality that shapes fault diagnosis here is threefold: cool-downs are
multi-day and costly, so downtime is expensive and fridge health sits on the critical path
of every experiment; hardware faults (leaks, blockages, quenches) are \emph{rare} and often
one-off, so labeled fault episodes are scarce; and each new fridge is commissioned with
\emph{no} fault history of its own. In practice, monitoring
is still built on threshold and rate-of-change alarms surfaced to a dashboard: they
answer \emph{is a channel out of band?} but not \emph{which physical fault is
developing and what should the operator do?} Closing that gap, turning telemetry
into a named diagnosis and a recommended action under real-world label scarcity, is the
problem we study.

We take the position that this is a natural task for an LLM \emph{agent}: the input
is a modest, heterogeneous, multi-channel time series; the output is a structured
judgment (detected / class / severity / action); and the reasoning benefits from
physical priors that a language model can be told in a prompt. But an agent is only
credible if it is measured against a strong classical baseline on realistic data.
This paper builds both sides of that comparison and reports the result honestly,
\emph{including where the agent loses}.

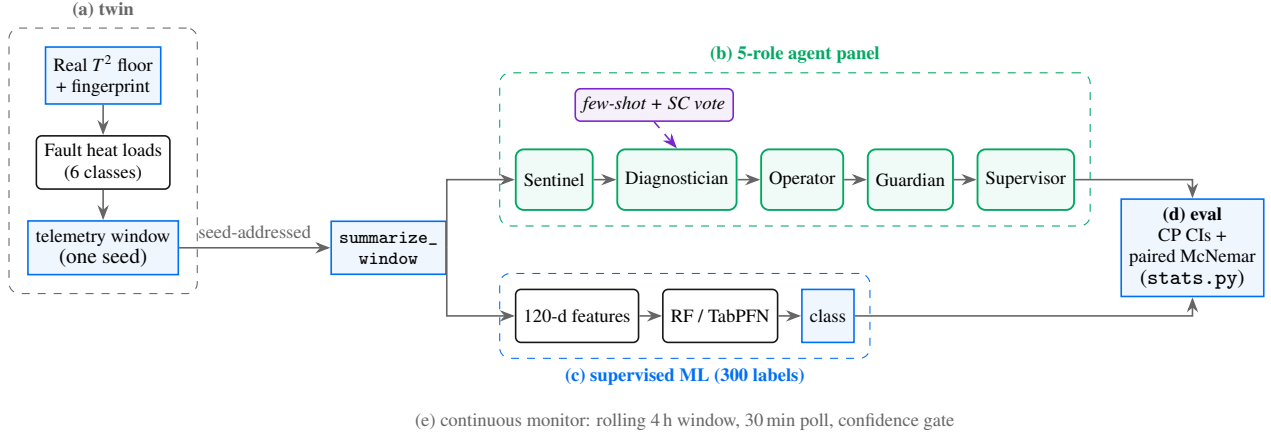
\begin{figure*}[t!]
\centering
\begin{tikzpicture}[node distance=3mm, every node/.style={font=\scriptsize}]
\node[data] (phys) {Real $T^2$ floor\\+ fingerprint};
\node[stage, below=4mm of phys] (fault) {Fault heat loads\\(6 classes)};
\node[data, below=4mm of fault] (win) {telemetry window\\{\footnotesize (one seed)}};
\draw[flow] (phys) -- (fault); \draw[flow] (fault) -- (win);
\begin{scope}[on background layer]
\node[draw=muted, dashed, rounded corners, fit=(phys)(fault)(win), inner sep=2.4mm,
  label={[muted,font=\scriptsize]above:\textbf{(a) twin}}] (twinbox) {};
\end{scope}
\node[data, right=20mm of win] (summ) {\texttt{summarize\_}\\\texttt{window}};
\draw[flow] (win) -- (summ)
  node[midway, above, font=\scriptsize, text=muted] {seed-addressed};
\node[agent, right=9mm of summ, yshift=9mm] (sen) {Sentinel};
\node[agent, right=3mm of sen] (dia) {Diagnostician};
\node[agent, right=3mm of dia] (op) {Operator};
\node[agent, right=3mm of op] (gu) {Guardian};
\node[agent, right=3mm of gu] (sup) {Supervisor};
\draw[flow] (sen)--(dia); \draw[flow] (dia)--(op); \draw[flow] (op)--(gu); \draw[flow] (gu)--(sup);
\node[lever, above=3.5mm of dia, xshift=-3mm] (lev) {few-shot + SC vote};
\draw[flow, oviolet, dashed] (lev.south) -- (dia.north);
\draw[flow] (summ.east) |- (sen.west);
\begin{scope}[on background layer]
\node[draw=ogreen, dashed, rounded corners, fit=(sen)(sup)(lev), inner sep=2mm,
  label={[ogreen,font=\scriptsize]above:\textbf{(b) 5-role agent panel}}] (panelbox) {};
\end{scope}
\node[stage, right=9mm of summ, yshift=-9mm] (feat) {120-d features};
\node[stage, right=3mm of feat] (rf) {RF / TabPFN};
\node[data, right=3mm of rf] (mlpred) {class};
\draw[flow] (feat)--(rf); \draw[flow] (rf)--(mlpred);
\draw[flow] (summ.east) |- (feat.west);
\begin{scope}[on background layer]
\node[draw=oblue, dashed, rounded corners, fit=(feat)(mlpred), inner sep=2mm,
  label={[oblue,font=\scriptsize]below:\textbf{(c) supervised ML (300 labels)}}] (mlbox) {};
\end{scope}
\node[data, right=6mm of sup, yshift=-9mm] (eval)
  {\textbf{(d) eval}\\ CP CIs +\\ paired McNemar\\{\footnotesize (\texttt{stats.py})}};
\draw[flow] (sup.east) -| (eval.north);
\draw[flow] (mlpred.east) -| (eval.south);
\node[muted, below=6mm of mlbox, font=\scriptsize] (mon)
  {(e) continuous monitor: rolling 4\,h window, 30\,min poll, confidence gate};
\end{tikzpicture}
\caption{End-to-end system. The twin (a) renders a telemetry window from one seed;
a single \texttt{summarize\_window} feeds the \emph{identical} window to both the 5-role
agent panel (b) and the supervised ML path (c). Because both methods see the same
seed-addressed scenarios, the head-to-head uses the exact paired McNemar test (d). The
same panel runs as a continuous monitor (e) over a rolling window at a fixed poll cadence.
Dashed violet: the optional few-shot/self-consistency levers (Section~\ref{sec:incontext}).}
\label{fig:system}
\end{figure*}

\smallskip
\noindent\textbf{Contributions:}
\begin{enumerate}
\item \textbf{A physics-grounded twin} (Section~\ref{sec:twin}) that couples a real
$T^2$ dilution floor, a fingerprint of real-fridge noise and cross-stage correlation,
and six physics-grounded fault classes, three of them engineered to be confusable on
temperature but separable on flow/pressure, so the benchmark is not trivially solved.
\item \textbf{A live 5-agent operations layer} (Section~\ref{sec:agents}) and a
1000-turn evaluation against a supervised ML opponent on \emph{identical} scenarios,
yielding an honest, non-overclaimed finding: zero-shot, the agent shows \emph{no
significant difference} on detection and \emph{loses} on classification, the gap tracing
to the label asymmetry (the ML model trains on 300 labels; the agent gets none).
\item \textbf{An evidence-based intervention} (Section~\ref{sec:incontext}): guided by
2025--2026 in-context-learning results, we add contrastive few-shot demonstrations and
self-consistency voting, and \emph{deliberately avoid} debate and self-refinement,
which recent work shows are dominated by compute cost. In a controlled comparison the
enhanced panel closes the entire classification gap. We do not claim these ICL mechanisms
as novel --- contrastive few-shot and self-consistency are established techniques. The
contribution is their disciplined application to a physics-constrained diagnostic task:
demonstrations curated by an \emph{a priori} thermal-degeneracy prior (Section~\ref{sec:incontext}),
an ablation that isolates which lever is load-bearing, and a negative-result methodology
(avoid debate/self-refine) grounded in 2026 evidence rather than momentum.
\item \textbf{A continuous-monitoring study} (Section~\ref{sec:monitor}): one unbroken
24\,h run in which the agent detects a developing helium leak $29.5$\,min after onset.
\item \textbf{Full reproducibility} (Section~\ref{sec:repro}): every figure and number
is generated from a released run artifact; nothing is hand-authored.
\end{enumerate}

\smallskip
\noindent\textbf{Takeaway.} On physically confusable cryogenic faults, a zero-shot agent
matches supervised ML on detection and, once given curated contrastive few-shot
demonstrations and self-consistency voting, closes the entire classification gap with no
parameter updates and six labeled examples; an ablation shows the multi-agent structure
buys \emph{auditable safety} (a separable Guardian veto and a per-turn action trail), not
extra accuracy. Beyond the specific result, the twin plus its seed-addressed harness are a
reusable, physics-constrained benchmark for LLM-agent diagnostic reasoning on
multi-channel instrument telemetry.

\section{Related Work}
\textbf{Cryogenic monitoring:} Facility-scale fridge monitoring today is threshold-
and lag-alert based (Grafana/Slack style). Machine-learning fault detection has been demonstrated on adjacent
large-scale cryogenic and superconducting systems: monitoring the LHC superconducting
magnets with LSTM networks~\cite{wielgosz}, model-learning anomaly detection in CERN
cryogenic control systems~\cite{tilaro}, anomaly detection and maintenance optimization for
large-scale cryogenic plants~\cite{cacace}, and SRF-cavity fault classification at Jefferson
Laboratory~\cite{tennant}. Complementary sensing modalities such as vibrational monitoring
at mK temperatures in dry dilution refrigerators are also emerging~\cite{bracevib}. Onnes
adds a diagnostic \emph{and} recommended-action layer on top of realistic dilution-fridge
telemetry, and pairs it with an LLM reasoning agent rather than a fixed ML model.

\textbf{Cryogenic infrastructure for quantum computing:} The dilution refrigerator is the
enabling platform for superconducting and spin qubits, and its microwatt-scale
mixing-chamber cooling budget is a first-order constraint on scaling. Recent work surveys
the enabling technologies for scalable superconducting quantum
computing~\cite{croot}, demonstrates wafer-scale packages containing $>$500
superconducting qubits within a single fridge~\cite{kennedy}, and operates digitally
controlled silicon spin-qubit processors at mK~\cite{hrl}. A major thrust is moving control
and readout into the cold, cryo-CMOS and hybrid photonic/CMOS controller
architectures~\cite{liuphotonic} and resource estimation of cryoelectronics for
fault-tolerant machines~\cite{kawabata}, all of which add heat load precisely where cooling
is scarcest. Cryogenic systems for quantum \emph{photonic} technologies face parallel
constraints~\cite{rubin}. This body of work motivates an operations layer that keeps such
increasingly loaded fridges healthy. On the agent side, LLMs are beginning to control
scientific instruments autonomously~\cite{xieauto}, the capability our operations layer
brings to cryogenic telemetry.

\textbf{In-context learning (ICL):} Few-shot prompting is a standard lever for
classification, but its behavior is nuanced. Many-shot ICL scales to hundreds of
examples on some tasks, yet 2026 work shows the established many-shot rules
\emph{break down for reasoning tasks}~\cite{manyshotcot}, and merely prompting an LLM
to ``reason over the series'' yields little gain in time-series
settings~\cite{timera}. It is the \emph{examples}, and their selection, that move the
needle. We therefore use \emph{curated} contrastive demonstrations rather than
many-shot dumping.

\textbf{Self-consistency vs.\ self-refinement:} Self-consistency, which samples a
solver $N$ times and majority-votes~\cite{selfconsistency}, is a cheap, reliable
classification lever. Recent controlled studies report it as the best accuracy-per-cost
operating point, while iterative self-refinement can \emph{reduce} accuracy
(e.g.\ $-4.6$ to $-9.1$ points in one 2026 study, against $+2.2$ for
self-consistency)~\cite{selfcorrect,madesign}. We adopt self-consistency and avoid
self-refine loops.

\textbf{Multi-agent debate:} Debate frameworks report reasoning gains, but 2026
analyses argue much of the apparent multi-agent advantage is an artifact of increased
test-time compute rather than coordination, once call budgets are
controlled~\cite{illusion}. Related work also documents belief-instability and
error-propagation failure modes in agent chains~\cite{delayedverify}. We keep a fixed,
auditable pipeline and add only mechanisms with strong cost-controlled evidence.

\textbf{LLM agents for fault diagnosis:} A fast-growing $2025$ literature applies LLM and
LLM-agent methods to fault diagnosis and predictive maintenance in industrial rotating
machinery, HVAC and building systems, power grids, and satellite/aerospace telemetry, and
LLMs are beginning to drive scientific instruments directly~\cite{xieauto}. Onnes differs on
two axes that this body of work largely leaves open. First, it is \emph{physics-grounded}:
faults are generated by a dilution-cooling forward model with a real-fridge noise
fingerprint, not by pattern-matching on unconstrained logs, so the confusable classes are
degenerate for a principled physical reason rather than by chance. Second, it targets
\emph{quantum-computing cryogenic infrastructure} --- a regime with microwatt cooling budgets,
mK operating points, and safety interlocks (never quench the magnet) that general industrial
PdM does not model. To our knowledge this is the first controlled agent-vs-supervised
head-to-head on dilution-refrigerator fault diagnosis.

\textbf{Tabular foundation models:} On small tabular problems, TabPFN and its
successors have shifted the frontier from gradient-boosted trees toward in-context
transformers~\cite{tabpfn25}, though gradient boosting remains highly competitive on
enterprise-scale tables~\cite{tabgap}. We use a random forest as the primary,
fully-reproducible opponent and discuss stronger tabular models in
Section~\ref{sec:discussion}.

\section{The Onnes Digital Twin}\label{sec:twin}
\begin{figure}[t]
\centering
\begin{tikzpicture}[node distance=3.2mm]
\node[data] (phys) {Real $T^2$ dilution floor\\(validated cooling law)};
\node[data, right=6mm of phys] (fp) {BlueFors fingerprint\\(noise + stage corr.)};
\node[stage, below=5mm of phys] (fault) {Physics-grounded\\fault heat loads};
\node[stage, below=5mm of fp] (imp) {Sensor imperfections\\(noise, dropout, rail)};
\node[data, below=5mm of $(fault)!0.5!(imp)$] (out)
  {FRTMS telemetry window\\{\footnotesize temp1--7, flow, $p_1$--$p_6$}};
\begin{scope}[on background layer]
\node[draw=muted, dashed, rounded corners, fit=(phys)(fp)(fault)(imp), inner sep=3mm,
  label={[muted,font=\scriptsize]above:\textbf{cryo\_engine.simulate (realistic path)}}] {};
\end{scope}
\draw[flow] (phys) -- (fault);
\draw[flow] (fp) -- (imp);
\draw[flow] (fault) -- (out);
\draw[flow] (imp) -- (out);
\draw[flow] (phys) -- (fp);
\end{tikzpicture}
\caption{Data flow of the twin. Physics sets the mean trajectory; a fingerprint
learned from real BlueFors logs supplies fluctuations and cross-stage correlation;
faults are heat-load perturbations; sensor imperfections are layered last.}
\label{fig:engine}
\end{figure}
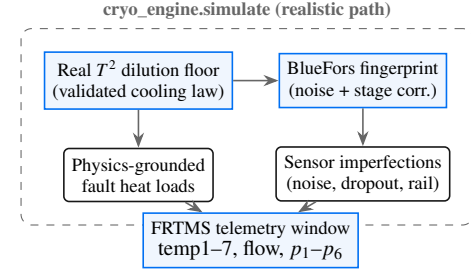

The twin emits the verified BlueFors/FRTMS telemetry schema (five stage temperatures
$50$\,K, $4$\,K, Still, cold plate, MXC; magnet channels; flow; pressures $p_1\!-\!p_6$)
so it is a drop-in for both the agent and ML pipelines (the end-to-end flow, including the
seed-addressed pairing that licenses the paired McNemar test, is shown in
Fig.~\ref{fig:system}; the twin, panel, and eval procedures are Algorithms~\ref{alg:twin}--\ref{alg:eval}).
It is built from three real ingredients (Fig.~\ref{fig:engine}).

\textbf{Physics mean:} Stage temperatures follow the $T^2$ dilution-cooling floor.
For a mixing-chamber heat load $\dot{Q}$, the dilution unit's cooling balance gives a
base temperature
\begin{equation}
T_{\mathrm{MXC}}(\dot{Q}) \;=\; \sqrt{\dot{Q}\,/\,\big[\dot{n}_3\,(95 - 11 f^2)\big]},
\label{eq:cool}
\end{equation}
where $\dot{n}_3$ is the ${}^3$He circulation rate and $f$ the concentration factor;
the cold plate follows the analogous $T_{\mathrm{CP}}=0.1\sqrt{\dot{Q}/\dot{Q}_{100}}$
anchored to the LD400 $100$\,mK spec. Crucially, cooling power vanishes as $T\!\to\!0$,
so there is a genuine base-temperature floor (unlike a linear model); the upper stages
($50$\,K, $4$\,K) are near-fixed by the pulse tube. Constants are tuned so an unloaded
MXC sits near $12$\,mK, consistent with a real BlueFors base state (numerically:
$\dot{n}_3{=}500\,\mu$mol/s, $f{=}1.5$, cold-plate $\dot{Q}_{100}{=}300\,\mu$W at $100$\,mK,
still cooling $30$\,mW; the $95/11$ enthalpy coefficients follow the LD400 calibration
$14\,\mu\mathrm{W}\!\to\!20$\,mK, all in \texttt{dilution\_cooling.py}); the functional form
is an engineering approximation sufficient for telemetry realism, not a first-principles
derivation of the mixture thermodynamics. The fingerprint is
estimated from public BlueFors dilution-fridge logs~\cite{cryometrics} (fridge
``blizzard,'' $2021$-$10$-$08$: channels CH1/2/5/6 temperature, flowmeter, and status,
released under \texttt{data/real/bluefors\_cryometrics\_sample}); per-stage relative noise
recovered from these logs (MXC $0.74\%$, $50$\,K $1.6\%$, flow $2.3\%$) is applied
multiplicatively (\S3, ``Real-fridge fingerprint''). We further
sanity-check stage base temperatures and cool-down behavior against two public
real-cryostat corpora, the University of Leeds DR-200 dilution-fridge cool-down
logs~\cite{leeds} and the ORNL Spallation Neutron Source cryogenic-moderator
dataset~\cite{ornlsns}.

\textbf{Real-fridge fingerprint:} Rather than toy Gaussian noise, per-stage relative
fluctuations and cross-stage correlations are learned from real BlueFors
logs~\cite{cryometrics} and applied multiplicatively to the physics mean, so a window ``looks like'' a real
machine. Concretely, a physics-mean channel $\bar{x}_c(t)$ is perturbed as
$x_c(t)=\bar{x}_c(t)\,\big(1+\varepsilon_c(t)\big)$ with $\boldsymbol{\varepsilon}(t)
\sim\mathcal{N}(\mathbf{0},\boldsymbol{\Sigma})$, where $\boldsymbol{\Sigma}$ is the
cross-stage covariance estimated from the logs.

\begin{figure}[t]
\centering
\includegraphics[width=\columnwidth]{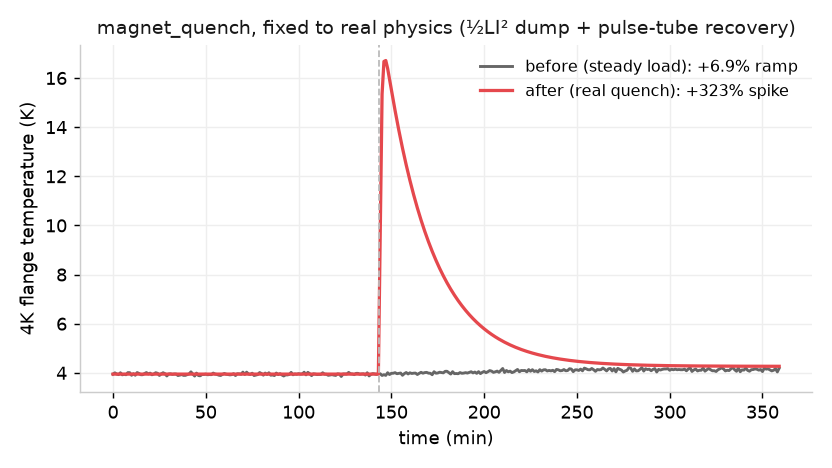}
\caption{Quench physics, corrected. Steady-load model: $+6.9\%$ ramp (grey).
Real $\tfrac12 L I^2$ transient: $+323\%$ spike then pulse-tube recovery (red).}
\label{fig:quench}
\end{figure}

\textbf{Fault classes:} Six labels drive the benchmark:
\texttt{normal}, \texttt{heat\_load\_spike}, \texttt{helium\_leak},
\texttt{magnet\_quench}, \texttt{wiring\_heat\_ingress}, \texttt{blocked\_impedance}.
Faults are injected as heat-load perturbations so temperatures come from the physics.
Two design choices make the benchmark non-trivial:

\emph{(i) A sharp quench transient:} A magnet quench dumps the stored magnetic energy
$E=\tfrac12 L I^2$ of the $9$\,T solenoid into the windings over seconds. We model the
$4$\,K-flange excursion as a difference-of-exponentials pulse (fast rise $\tau_r$, slow
recovery $\tau_d$),
\begin{equation}
\Delta T_{4\mathrm{K}}(t) = A\,s\,\big(e^{-(t-t_0)/\tau_d}-e^{-(t-t_0)/\tau_r}\big)_+
  \; + \; \text{residual},
\label{eq:quench}
\end{equation}
with $\tau_r\!\approx\!0.8$\,min, $\tau_d\!\approx\!25$\,min, and severity $s$.
Fig.~\ref{fig:quench} contrasts the fixed physics: a steady-load model produces only a
gentle $+6.9\%$ ramp on the $4$\,K flange, whereas the corrected quench produces a
$+323\%$ spike followed by pulse-tube recovery.

\begin{figure}[t]
\centering
\includegraphics[width=\columnwidth]{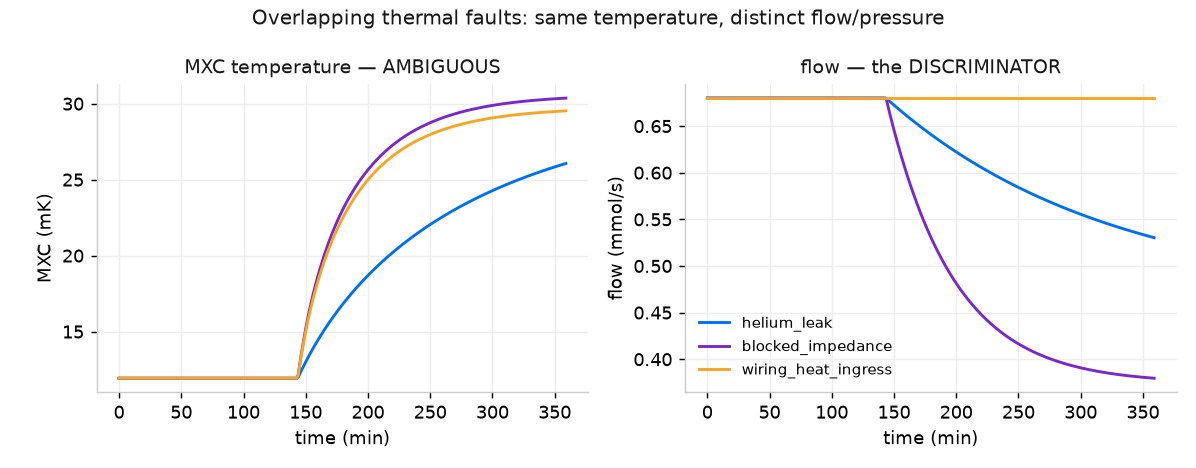}
\caption{Three thermal faults overlap on MXC temperature (left, ambiguous) but
separate on the flow channel (right, discriminative). A static temperature snapshot
cannot classify them; a supervised model can still reach macro-$F_1\;0.983$ from
temperature \emph{trajectory} features alone (Table~\ref{tab:baselines}), so the
overlap is a difficulty for snapshot- and zero-shot reasoning, not an
information-theoretic barrier. Flow and pressure are the physically direct, noise-robust
discriminators, and are what the engineered baseline (Section~\ref{sec:ml}) exploits.}
\label{fig:overlap}
\end{figure}

\emph{(ii) Deliberately confusable thermal faults:} \texttt{helium\_leak},
\texttt{blocked\_impedance}, and \texttt{wiring\_heat\_ingress} are tuned to
\emph{overlap on temperature} while \emph{separating on flow and pressure}
(Fig.~\ref{fig:overlap}). This is what makes classification (not just detection) the
hard part, and it is where the zero-shot agent later fails.

\section{The Multi-Agent Operations Layer}\label{sec:agents}
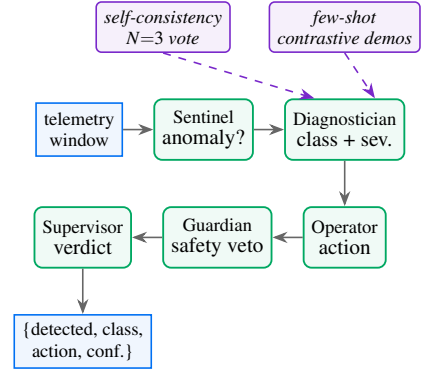
\begin{figure}[t]
\centering
\begin{tikzpicture}[node distance=3.0mm]
\node[data] (win) {telemetry\\window};
\node[agent, right=4mm of win] (sen) {Sentinel\\{\footnotesize anomaly?}};
\node[agent, right=4mm of sen] (dia) {Diagnostician\\{\footnotesize class + sev.}};
\node[agent, below=6mm of dia] (op) {Operator\\{\footnotesize action}};
\node[agent, left=4mm of op] (gu) {Guardian\\{\footnotesize safety veto}};
\node[agent, left=4mm of gu] (sup) {Supervisor\\{\footnotesize verdict}};
\node[data, below=6mm of sup] (out) {\{detected, class,\\action, conf.\}};
\draw[flow] (win) -- (sen);
\draw[flow] (sen) -- (dia);
\draw[flow] (dia) -- (op);
\draw[flow] (op) -- (gu);
\draw[flow] (gu) -- (sup);
\draw[flow] (sup) -- (out);
\node[lever, above=6mm of dia] (fs) {few-shot\\contrastive demos};
\node[lever, left=5mm of fs] (sc) {self-consistency\\$N{=}3$ vote};
\draw[flow, oviolet, dashed] (fs.south) -- ([xshift=4mm]dia.north);
\draw[flow, oviolet, dashed] (sc.south) -- ([xshift=-4mm]dia.north);
\end{tikzpicture}
\caption{The five-role pipeline. Solid: the fixed zero-shot path. Dashed (violet):
the two optional, evidence-backed levers attached to the Diagnostician
(Section~\ref{sec:incontext}). Defaults preserve exact zero-shot behavior.}
\label{fig:pipeline}
\end{figure}

\begin{figure*}[t]
\centering
\includegraphics[width=\textwidth]{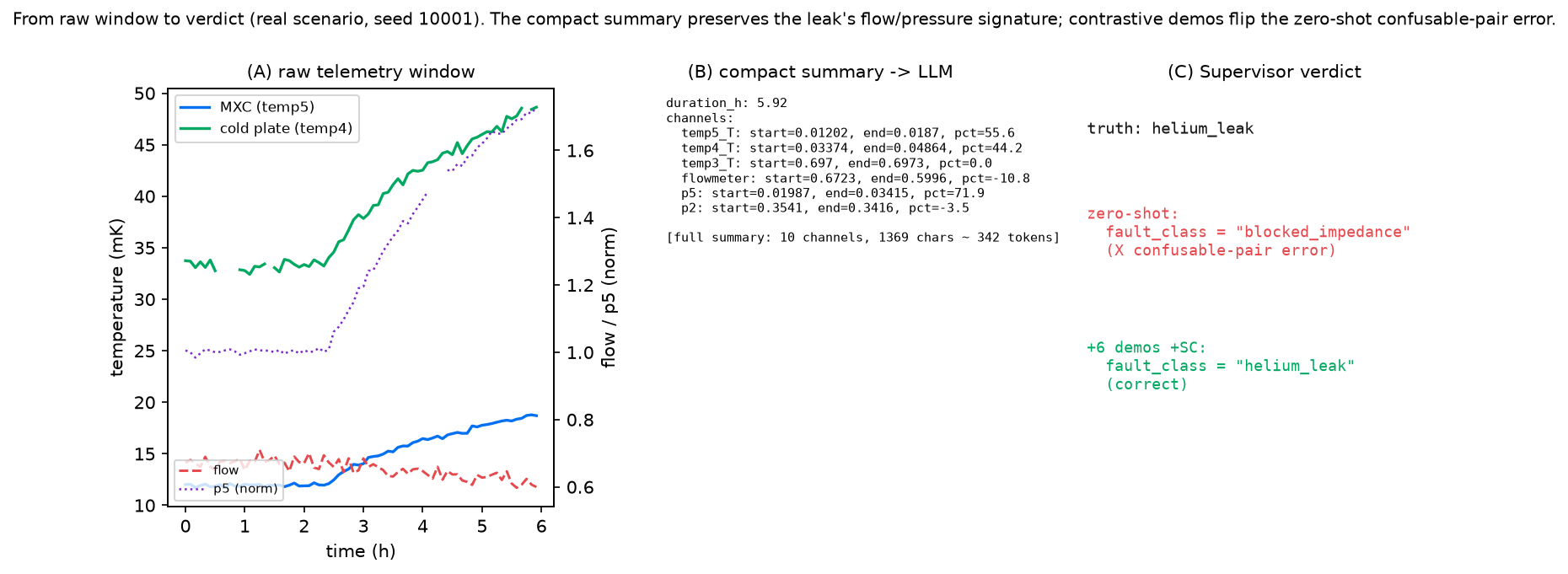}
\caption{From raw window to verdict for one real scenario (eval seed $10001$, truth
\texttt{helium\_leak}). (A) the discriminating channels: MXC/cold-plate warm while flow
drops and the OVC gauge $p_5$ spikes --- the leak signature. (B) the exact compact summary
fed to the LLM ($\sim\!340$ tokens), which preserves the flow/pressure contrast but coarsens
the trajectory. (C) the actual logged Supervisor verdicts: zero-shot commits the
confusable-pair error (\texttt{blocked\_impedance}); the enhanced panel recovers
\texttt{helium\_leak}.}
\label{fig:window}
\end{figure*}

The operations layer is a fixed, auditable five-role pipeline over one telemetry
window (Fig.~\ref{fig:pipeline}). Each role is a single LLM call. Our primary backend
is Claude Opus~4.8 (queried through an OpenAI-compatible proxy); to test whether the
findings are model-specific we replicate the entire evaluation on Google Gemini~3.1~Pro
(\texttt{gemini-3.1-pro-preview}, via the Google GenAI SDK), reusing the identical
scenarios, prompts, parser, and scorer so the model is the only variable
(Section~\ref{sec:discussion}). The window is presented as a compact numeric summary
(per-channel start/end/\%-change plus a coarse trajectory), not raw rows, mirroring a
realistic context budget; Fig.~\ref{fig:window} shows a real window, its summary, and the
resulting verdicts side by side.

\textbf{Roles:} \emph{Sentinel} flags whether an anomaly is developing;
\emph{Diagnostician} assigns the fault class and severity; \emph{Operator} proposes one
corrective action; \emph{Guardian} vetoes unsafe actions (e.g.\ anything that could
quench the magnet); \emph{Supervisor} reconciles the panel into a final verdict
(the full call sequence, with the optional levers, is Algorithm~\ref{alg:panel}).
Thus $N$ scenarios yield $5N$ agent turns; $N{=}200$ yields $1000$ turns. Every turn
(role, prompt summary, raw JSON reply, latency) is logged to JSONL for audit.

\textbf{Why five roles, not one prompt?} We state the design thesis up front because it,
not accuracy, is what the decomposition buys --- and our own ablation
(Section~\ref{sec:incontext}) will show a single well-prompted call \emph{matches} the
panel's classification accuracy. The panel's value is instead operational, on the axis that
matters when an autonomous agent can command real hardware: (i)~the roles expose
\emph{independent operating points}, so a deliberately high-recall Sentinel can be gated by a
conservative Supervisor --- catching developing faults early without paying the false-alarm
cost at the final verdict, a trade-off a monolithic call cannot expose; (ii)~the Guardian is
a \emph{single inspectable veto surface} on a code path independent of the (possibly wrong)
classification, so a misdiagnosis cannot silently become a magnet-quenching command;
(iii)~the per-role JSONL log is a \emph{post-incident audit trail} that localizes \emph{which}
role failed, where a single call collapses the decision into one opaque output. This
separable-safety framing is what distinguishes the panel from accuracy-chasing multi-agent
stacks; we quantify the Guardian's live veto rate in Section~\ref{sec:headtohead} and give
the full argument in Section~\ref{sec:discussion}.

\textbf{Evaluation protocol:} Scenarios are seed-addressed so the agent and the ML
opponent see \emph{identical} telemetry. Difficulty is raised with the benchmark's
realism stressor (severity scaled by $0.5$) so the ML model is not trivially perfect.
We score \emph{detection} (normal vs.\ any-fault, from the Supervisor's explicit flag)
and \emph{classification} (exact fault class), using the same metric code as the ML
benchmark so the numbers are directly comparable.

\textbf{Metrics and statistics:} Detection is summarized by $F_1=2PR/(P{+}R)$;
multi-class performance by accuracy and macro-$F_1=\frac{1}{|C|}\sum_{c\in C}F_1^{(c)}$
over the present classes $C$. Every proportion is reported with an exact
Clopper--Pearson~\cite{clopperpearson1934} $95\%$ interval: for $k$ successes in $n$ trials the bounds are the
Beta quantiles $[\,B_{\alpha/2}(k,n{-}k{+}1),\,B_{1-\alpha/2}(k{+}1,n{-}k)\,]$. Two
classifiers on the \emph{same} scenarios are compared with the exact paired McNemar~\cite{mcnemar1947}
test: with $b,c$ the discordant counts (one right where the other errs), the two-sided
$p$-value is the binomial tail $2\!\sum_{i\le\min(b,c)}\binom{b+c}{i}2^{-(b+c)}$. These
are computed from the released turn logs by \texttt{onnesim/stats.py}.

\section{The Supervised ML Opponent}\label{sec:ml}
\begin{figure}[t]
\centering
\includegraphics[width=\columnwidth]{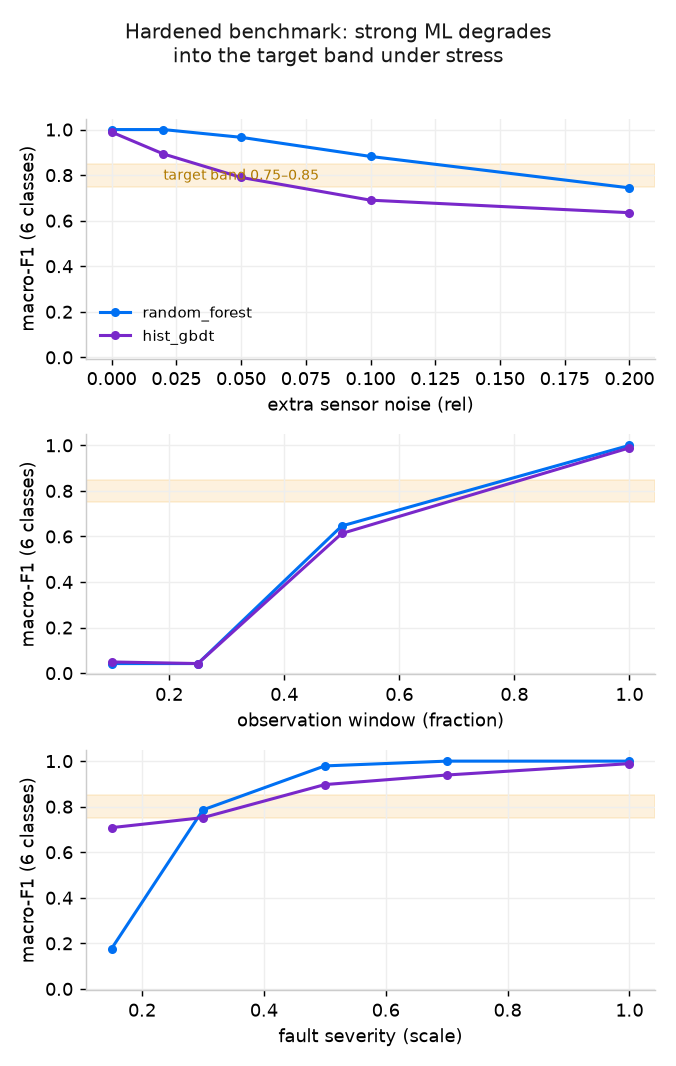}
\caption{Supervised ML under realism stress. Macro-$F_1$ degrades as the observation
window shrinks and severity drops, into a non-saturated regime.}
\label{fig:stress}
\end{figure}

The opponent is a two-stage classifier over a 120-dimensional feature vector
(per-channel statistics and trajectory features): a detector (normal vs.\ fault) and a
fault-type classifier. The primary model is a random forest~\cite{breiman2001} trained on
$300$ \emph{clean} realistic scenarios drawn from a seed range disjoint from the
evaluation set. Under the realism stress test (Fig.~\ref{fig:stress}), the model is
strong at full information but degrades into a target band as the observation window
shrinks (macro-$F_1$ falling from $1.00$ at full window to $0.646$ at half window),
confirming the benchmark is not saturated. Its residual confusions under sensor noise
lie on the same physically-adjacent faults (helium leak / wiring ingress / blocked
impedance) that the benchmark engineers to overlap.

\textbf{A baseline zoo, not a straw man:} To ensure the opponent is competitive, we
evaluate a baseline zoo (random forest, histogram gradient boosting, LightGBM~\cite{lightgbm}, and the
TabPFN-2.5 tabular foundation model~\cite{tabpfn25}) under one seed-addressed protocol on
the \emph{same} $n{=}200$ eval seeds the agent faces (Table~\ref{tab:baselines}).
TabPFN-2.5 is the strongest (macro-$F_1\;0.996$), edging the random forest ($0.983$),
with gradient boosting behind; we therefore report the head-to-head against the random
forest as the primary, fully-local opponent and against TabPFN-2.5 as the stronger
foundation-model bar. Either way the supervised ceiling is high, which is exactly what
makes the zero-shot classification gap, and its later closing, meaningful.

\textbf{An engineered physics-rule baseline (what an operator builds first):} A learned
model is not the only non-LLM reference. A skilled cryogenic engineer would not reach for
a classifier first --- they would write rules: rate-of-change on the cold stages plus the
sign of $\Delta$flow and $\Delta p$ on the channels that physically separate the
overlapping faults (a helium leak drops circulation flow and raises the OVC gauge $p_5$; a
blocked impedance drops flow but raises the condenser gauge $p_1$; wiring ingress leaves
flow and pressure flat). We implement exactly this (\texttt{evaluate.engineered\_baseline},
using \emph{no} training labels, with thresholds set a few $\sigma$ above the measured
sensor-noise floors) and score it on the identical $n{=}200$ held-out seeds. It reaches
multiclass accuracy $\ENGmulti$ (macro-$F_1\;\ENGmacro$) --- far above the fixed-cutoff
FRTMS threshold rules ($\ENGthreshMulti$) and close to the random forest ($0.985$). This
sharpens rather than weakens the case for a learned/agent system in two ways. First, it
shows how much of the task is solved by good physics engineering alone, so the agent's and
RF's remaining $\sim\!7$ points are honestly scoped. Second, the rule system is
\emph{brittle}: under matched sensor noise its macro-$F_1$ collapses from $\ENGmacro$
(clean) to $\ENGnoiseEng$ at $20\%$ noise, where the random forest still holds $\ENGnoiseRF$
and the agent reasons from the same noisy summary. Hand-tuned thresholds do not survive the
real-fridge noise the fingerprint injects; a learned or reasoning system degrades
gracefully. Verbatim numbers in \texttt{engineered\_baseline\_h2h.json}.

\begin{table}[t]
\caption{Baseline zoo on identical $n{=}200$ held-out seeds ($\mathrm{sev}\times0.5$),
same features and protocol as the head-to-head. Clopper--Pearson $95\%$ CIs. Learned
models verbatim from \texttt{baseline\_zoo.json}; the two rule baselines (no training
labels) from \texttt{engineered\_baseline\_h2h.json}. TabPFN-2.5 is the strongest opponent;
the engineered physics rules recover most of the task, while the fixed-cutoff FRTMS rules
do not. \emph{These near-saturated scores measure simulator separability, not
generalization to hardware}: the fault signatures are authored at high SNR (e.g.\ a
\texttt{helium\_leak} drives the OVC gauge $p_5$ up to $+300\%$ against a real-log pressure-noise
floor of $0.6\%$, and flow down $35\%$ against a $2.3\%$ floor), so a forest separates them
near-perfectly. The benchmark's value is the \emph{confusable-pair contrast structure}
(Fig.~\ref{fig:overlap}), not the absolute accuracy --- as the source itself flags
(``$F_1\,0.997$\dots a red flag, not a trophy'').}
\label{tab:baselines}
\centering
\setlength{\tabcolsep}{4pt}
\resizebox{\columnwidth}{!}{%
\begin{tabular}{lcc}
\toprule
\textbf{Model} & \textbf{Macro-$F_1$} & \textbf{Class.\ acc.\ [95\% CI]}\\
\midrule
TabPFN-2.5 & $\mathbf{0.996}$ & $0.995$ {\footnotesize[.97,1.0]}\\
Random forest & $0.983$ & $0.985$ {\footnotesize[.96,1.0]}\\
Hist.\ grad.\ boosting & $0.883$ & $0.890$ {\footnotesize[.84,.93]}\\
LightGBM & $0.860$ & $0.865$ {\footnotesize[.81,.91]}\\
Logistic regression & $0.843$ & $0.825$ {\footnotesize[.77,.88]}\\
\midrule
Engineered rules {\footnotesize(0 labels)} & $\ENGmacro$ & $\ENGmulti$ {\footnotesize[.87,.95]}\\
Threshold rules {\footnotesize(FRTMS)} & $0.173$ & $\ENGthreshMulti$ {\footnotesize[.19,.31]}\\
\bottomrule
\end{tabular}}
\end{table}

\begin{figure}[t]
\centering
\includegraphics[width=0.95\columnwidth]{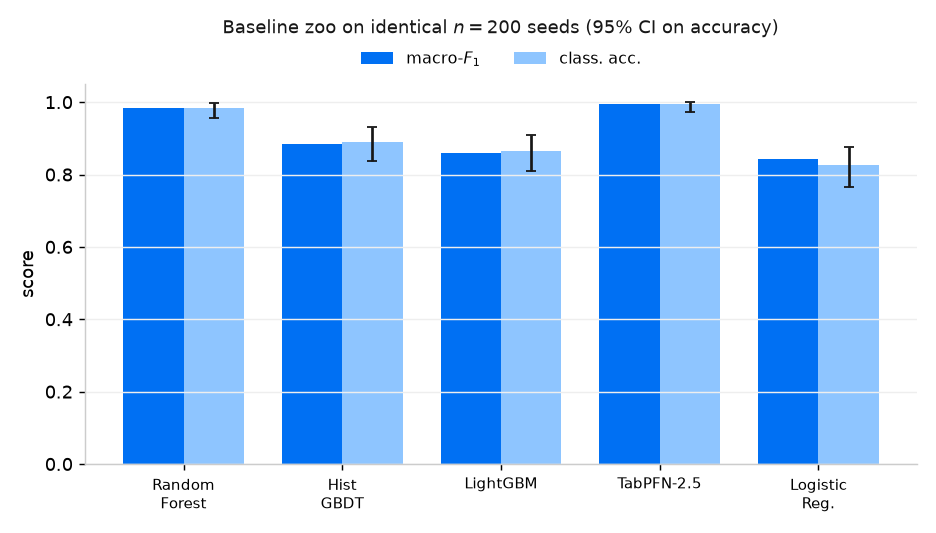}
\caption{Baseline zoo on identical $n{=}200$ held-out seeds ($\mathrm{sev}\times0.5$),
with $95\%$ CIs on accuracy. TabPFN-2.5 is the strongest opponent; we keep the random
forest as the primary head-to-head opponent because it is fully local and
reproducible, and report TabPFN-2.5 as the stronger foundation-model bar.}
\label{fig:zoo}
\end{figure}

\begin{table}[t]
\caption{Measured cost per telemetry window. Agent latencies are per-role means from the
run logs; RF timings are direct measurements. Verbatim from \texttt{cost\_model.json}.
The agent's value is reasoning from few or no labeled examples, not speed.}
\label{tab:cost}
\centering
\resizebox{\columnwidth}{!}{%
\begin{tabular}{lcc}
\toprule
\textbf{Method} & \textbf{LLM calls/window} & \textbf{Latency}\\
\midrule
Agent panel (zero-shot) & $5$ & $\sim\!88$\,s\\
Agent panel ($+$ vote $N{=}3$) & $7$ & $\sim\!112$\,s\\
Single agent (ablation) & $1$ & $\sim\!18$\,s\\
Random forest & $0$ & $0.18$\,ms\\
\bottomrule
\end{tabular}%
}
\end{table}

\section{Head-to-Head: Zero-Shot Agent vs.\ Supervised ML}\label{sec:headtohead}
Table~\ref{tab:headtohead} reports the full 1000-turn evaluation
($200$ scenarios). The result is deliberately non-overclaimed.

\begin{table}[t]
\caption{Zero-shot agent panel vs.\ supervised ML, $n{=}200$ identical held-out
scenarios ($1000$ agent turns). Clopper--Pearson $95\%$ CIs in brackets; the last
column is the exact paired McNemar test between the two methods. Numbers verbatim from
\texttt{agent\_eval\_results.json} and \texttt{head\_to\_head\_stats.json}.}
\label{tab:headtohead}
\centering
\small
\setlength{\tabcolsep}{4pt}
\resizebox{\columnwidth}{!}{%
\begin{tabular}{lccc}
\toprule
& \textbf{Detection $F_1$} & \textbf{Classification acc.} & \textbf{McNemar}\\
\midrule
Zero-shot agent & $0.979$ & $0.685$ {\footnotesize[.62,.75]} & ---\\
Supervised RF & $0.997$ & $0.985$ {\footnotesize[.96,1.0]} & ---\\
\midrule
\emph{detection} & \multicolumn{2}{c}{agent vs.\ RF} & $p=0.07$\\
\emph{classification} & \multicolumn{2}{c}{agent vs.\ RF} & $p<10^{-15}$\\
\bottomrule
\end{tabular}%
}
\end{table}

\begin{figure}[t]
\centering
\includegraphics[width=\columnwidth]{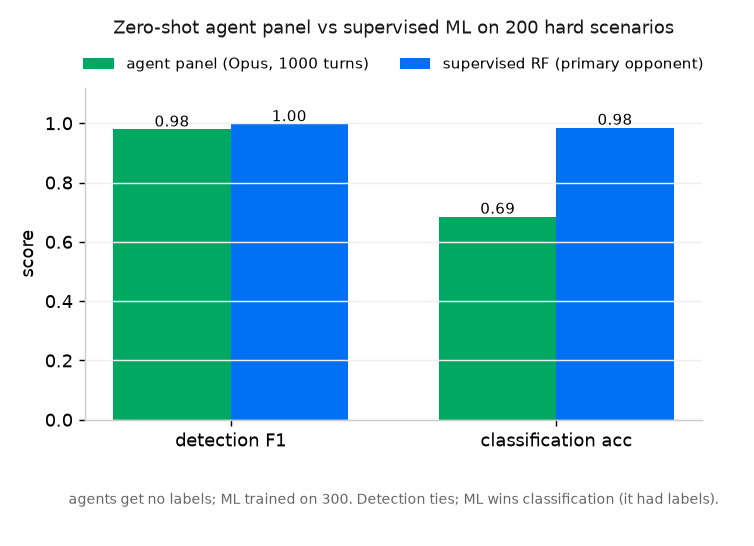}
\caption{Head-to-head. Detection shows no significant difference; the supervised model
wins classification, having trained on 300 labels the zero-shot agent never saw.}
\label{fig:avm}
\end{figure}

\begin{figure}[t]
\centering
\includegraphics[width=\columnwidth]{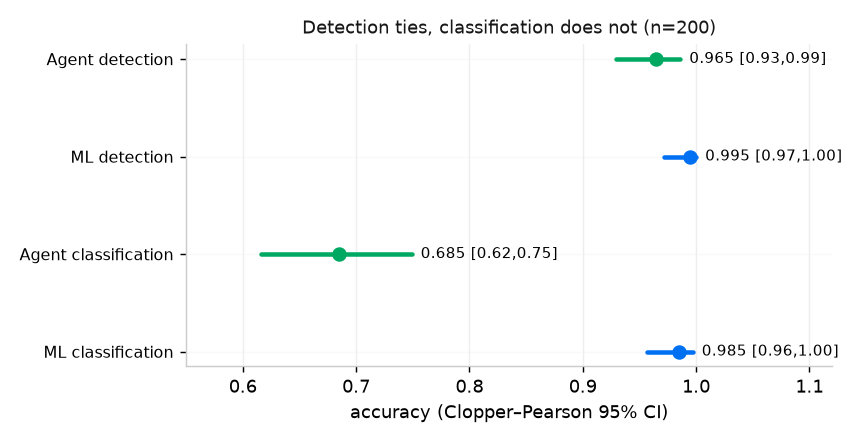}
\caption{Clopper--Pearson $95\%$ confidence intervals on \emph{accuracy} ($n{=}200$;
detection accuracy $0.965$/$0.995$, classification accuracy $0.685$/$0.985$ for
agent/RF). These are the raw-accuracy proportions the exact tests run on, and differ
from the detection $F_1$ reported in Table~\ref{tab:headtohead}. The detection
intervals overlap heavily (McNemar $p=0.07$: no significant difference), while the classification
intervals are disjoint (McNemar $p<10^{-15}$: the ML advantage is not sampling noise).}
\label{fig:ciforest}
\end{figure}

\begin{figure}[t]
\centering
\includegraphics[width=0.86\columnwidth]{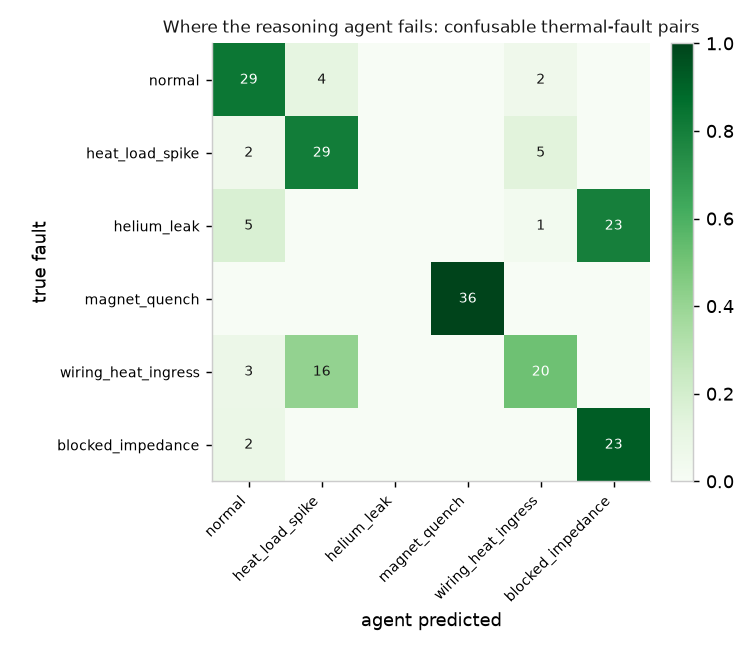}
\caption{Zero-shot agent confusion matrix on the $n{=}200$ held-out scenarios (rows: true
class; columns: the Supervisor's predicted class; cells: scenario counts). The two dominant
off-diagonal cells are the engineered confusable pairs ---
\texttt{helium\_leak}$\rightarrow$\texttt{blocked\_impedance} ($23$ cases) and
\texttt{wiring\_heat\_ingress}$\rightarrow$\texttt{heat\_load\_spike} ($16$ cases) --- which
the twin was built to make ambiguous on temperature; the off-diagonal mass therefore
concentrates exactly where a priori physics predicts, evidence the evaluation measures what
it was designed to rather than diffuse error. The enhanced panel closes these cells
(Fig.~\ref{fig:confpanels}, Table~\ref{tab:perclass}).}
\label{fig:agentconf}
\end{figure}

\begin{figure*}[t]
\centering
\includegraphics[width=\textwidth]{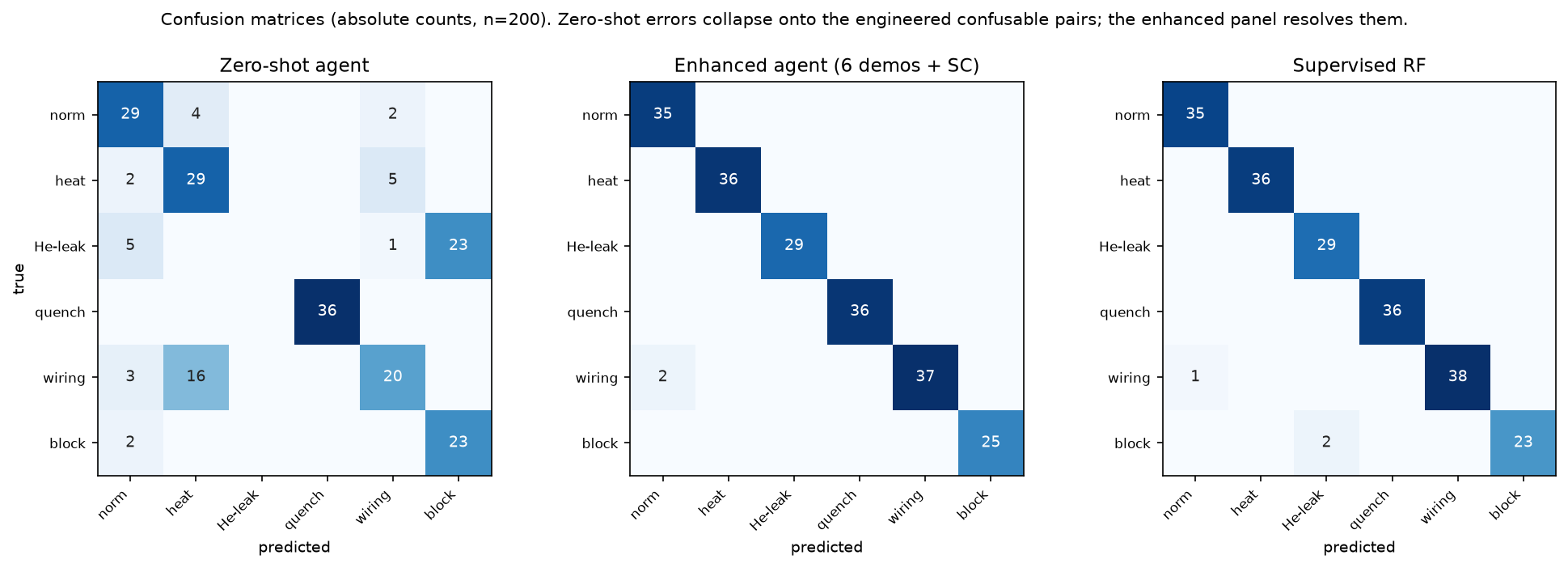}
\caption{Confusion matrices with absolute counts ($n{=}200$; recomputed from the released
turn logs by \texttt{scripts/analyze\_turn\_logs.py}). Zero-shot errors collapse almost
entirely onto the two engineered confusable pairs --- $23/29$ \texttt{helium\_leak} windows
are called \texttt{blocked\_impedance}, and $16$ \texttt{wiring\_heat\_ingress} are called
\texttt{heat\_load\_spike}. The enhanced panel and RF are near-diagonal.}
\label{fig:confpanels}
\end{figure*}

\textbf{Detection: no significant difference; classification: a real gap.} On
\emph{detection}, the zero-shot panel is within $0.02$ of the ML model ($0.979$ vs.\
$0.997$): asked only ``is something wrong?'', reasoning from physics priors suffices,
and the exact paired McNemar test finds no significant difference ($p=0.07$; only $8$
discordant scenarios). We report this as ``no significant difference detected,'' not as
proven equivalence: with $n{=}200$ the test simply lacks power to certify a gap this
small, and a formal equivalence test (TOST) within a pre-specified margin is future work.
On \emph{classification}, the panel trails ($0.685$ vs.\ $0.985$, $p<10^{-15}$), and this
gap is unambiguous. Crucially, the errors are not random:
Fig.~\ref{fig:agentconf} shows they concentrate on the engineered confusable pairs
(\texttt{helium\_leak}$\rightarrow$\texttt{blocked\_impedance}, 23 cases;
\texttt{wiring\_heat\_ingress}$\rightarrow$\texttt{heat\_load\_spike}, 16 cases), with a
smaller scattered tail elsewhere. The
agent reasons about the right channels but selects the wrong twin, exactly where
temperature is ambiguous (Fig.~\ref{fig:overlap}). Counts for all three conditions appear
in Fig.~\ref{fig:confpanels}, and the per-class breakdown (Table~\ref{tab:perclass}) shows
the zero-shot gap is entirely on the confusable classes (\texttt{helium\_leak} $F_1{=}0.000$),
which the enhanced panel closes.

\begin{table}[t]
\caption{Per-class $F_1$ (support $n$) on the identical $n{=}200$ held-out scenarios,
recomputed from the turn logs. The zero-shot gap is not diffuse: it is concentrated on the
engineered confusable classes (\texttt{helium\_leak} collapses to $0.000$; every one is
mislabeled as its \texttt{blocked\_impedance} twin), while \texttt{magnet\_quench} is
already perfect. Curated demonstrations $+$ self-consistency close exactly those cells.}
\label{tab:perclass}
\centering
\resizebox{\columnwidth}{!}{%
\begin{tabular}{lcccc}
\toprule
\textbf{Class} & $n$ & \textbf{Zero-shot} & \textbf{Enhanced} & \textbf{RF}\\
\midrule
\texttt{normal} & $35$ & $0.763$ & $0.972$ & $0.986$\\
\texttt{heat\_load\_spike} & $36$ & $0.682$ & $1.000$ & $1.000$\\
\texttt{helium\_leak} & $29$ & $\mathbf{0.000}$ & $1.000$ & $0.967$\\
\texttt{magnet\_quench} & $36$ & $1.000$ & $1.000$ & $1.000$\\
\texttt{wiring\_heat\_ingress} & $39$ & $0.597$ & $0.974$ & $0.987$\\
\texttt{blocked\_impedance} & $25$ & $0.648$ & $1.000$ & $0.958$\\
\midrule
macro-$F_1$ & & $0.615$ & $0.991$ & $0.983$\\
\bottomrule
\end{tabular}%
}
\end{table}

\textbf{Guardian veto rate.} The multi-agent structure's value is a separable, auditable
safety surface, so we quantify how often it fires: recomputed from the released turn logs,
the Guardian returned a parseable verdict on $185/200$ zero-shot windows and vetoed the
Operator's proposed action on $14$ of them ($7.6\%$; $9.0\%$, $17/189$, on the enhanced
run). The veto is thus a live, non-trivial choke point rather than a rubber stamp ---
consistent with treating the panel's contribution as auditable safety rather than accuracy
(Section~\ref{sec:discussion}).

\textbf{Honest reading:} This is a fair-but-asymmetric comparison: the ML model is
\emph{supervised} (300 labels) and the agent is \emph{zero-shot} (a prompt, no
labels). That detection shows no significant difference zero-shot, and that the failures
are interpretable and physically localized, is the real result, not ``ML beats agents.''
In a real fridge, faults are rare and labels scarce, so a method that reaches useful
accuracy from few or no diagnosed episodes has genuine operational value. This asymmetry
motivates the intervention below.

\section{Closing the Gap with In-Context Techniques}\label{sec:incontext}
If the agent's only weakness is naming confusable classes, and the ML model's only
advantage is labeled data, then in-context examples should help \emph{without} training.
We add two mechanisms to the Diagnostician (Fig.~\ref{fig:pipeline}, violet), both
chosen for strong, cost-controlled evidence:

\textbf{(1) Contrastive few-shot demonstrations:} We construct $k{=}6$ labeled
example windows drawn from a seed range \emph{disjoint} from both the ML training seeds
and the evaluation seeds (no test leakage), over-weighting the confusable pairs, and
prepend them to the Diagnostician prompt in the same compact-summary format the agent
sees at inference. We use curated few-shot rather than many-shot because many-shot ICL
degrades on reasoning tasks~\cite{manyshotcot}. The decision to over-weight the confusable
pairs uses only \emph{a priori} physics, not held-out test labels: the thermal degeneracy
is predictable before running the benchmark, because \texttt{helium\_leak},
\texttt{blocked\_impedance}, and \texttt{wiring\_heat\_ingress} all raise the same cold
stages and are separable only on flow and pressure (Section~\ref{sec:twin},
Fig.~\ref{fig:overlap}). Any cryogenic operator would anticipate this ambiguity from the
fault mechanisms alone; the curation encodes that domain prior, and the demonstration
seeds never intersect the evaluation set.

\textbf{(2) Self-consistency voting:} We sample the Diagnostician $N{=}3$ times at a
diversity temperature and majority-vote the fault class~\cite{selfconsistency}. We
\emph{do not} use self-refinement or multi-round debate, following evidence that the
former hurts accuracy~\cite{selfcorrect} and the latter's gains are largely a
test-time-compute artifact~\cite{illusion}.

Both levers are backward-compatible: with $k{=}0,\,N{=}1$ the pipeline is byte-identical
to the zero-shot baseline, so the baseline reproduces exactly.

\textbf{Controlled result:} We evaluate the enhanced panel on held-out scenarios and
compare against the zero-shot panel and the supervised model on the \emph{same}
scenarios (Table~\ref{tab:lift}). On the $n{=}24$ proof-of-concept set the two
techniques raise classification accuracy from $0.50$ to $1.00$, a $+0.50$ absolute
lift, matching the supervised model with no task-specific training. Every zero-shot
error on this set was a confusable-pair confusion; the contrastive demonstrations and
vote resolved exactly those. We then confirm the effect at scale: on the \emph{same}
$n{=}200$ seeds as the zero-shot anchor (Table~\ref{tab:headtohead}), the enhanced
panel raises classification from $0.685$ to \CONFTWOHUNDRED. A paired McNemar test
gives $p<10^{-16}$ ($62$ scenarios fixed, $1$ regressed), so the lift is a large-$n$,
statistically decisive effect, not an artifact of the small set. At $0.990$ the
enhanced panel reaches parity with the supervised random forest
($0.985$); the $0.005$ gap is a single scenario at $n{=}200$ and we make no claim of
superiority, only that a training-free method (no parameter updates; six labeled
demonstrations) closes the zero-shot classification gap
entirely. It also lands within $0.006$ of the strongest supervised baseline in our zoo,
TabPFN-2.5 at $0.996$ (Table~\ref{tab:baselines}), under a fully reproducible local
protocol. Detection $F_1$ also edges up under the levers ($0.979\!\to\!0.994$; see
Table~\ref{tab:backend}): although the levers attach only to the Diagnostician, the
Supervisor conditions its final detected/not-detected flag on the Diagnostician's
output, so a sharper class call also sharpens the binary detection decision.

\begin{table}[t]
\caption{Controlled comparison on \emph{identical} held-out scenarios. Numbers verbatim
from \texttt{technique\_lift.json} ($n{=}24$) and
\texttt{agent\_eval\_fewshot\_n200\_results.json} ($n{=}200$). The zero-shot and
enhanced agents differ only by the two levers of Section~\ref{sec:incontext}.}
\label{tab:lift}
\centering
\begin{tabular}{lcc}
\toprule
\textbf{Condition} & \textbf{$n{=}24$} & \textbf{$n{=}200$}\\
\midrule
Zero-shot agent (prompt only) & $0.50$ & $0.685$\\
\;\;+ few-shot + self-consistency & $\mathbf{1.00}$ & $\mathbf{\CONFTWOHUNDREDSHORT}$\\
Supervised random forest & $1.00$ & $0.985$\\
\bottomrule
\end{tabular}
\end{table}

\textbf{Scope:} The $n{=}24$ comparison is a proof-of-concept; the $24$ scenarios are
the held-out draws on which the zero-shot panel scored $0.50$ (they were not selected
to favor the enhancement, and the levers were fixed beforehand). The $n{=}200$ column,
on the identical seeds as the $1000$-turn zero-shot anchor
(Table~\ref{tab:headtohead}), is the large-$n$ confirmation; we report both and label
their sizes explicitly.

\textbf{Ablation:} Because the enhancement adds two levers at once, and because a
five-role panel is more expensive than one call, we ablate both factors on a common
held-out set (Table~\ref{tab:ablation}): each lever alone (few-shot only,
self-consistency only) versus both, and the five-role panel versus a single-call agent
under matched levers. The single-vs-panel row directly tests the concern behind our
fixed pipeline: that multi-agent structure may add cost without accuracy over
chain-of-thought with self-consistency~\cite{illusion}. The results
(Table~\ref{tab:ablation}) attribute the lift almost entirely to curated few-shot: it
alone lifts the panel from $0.500$ to $0.983$, while self-consistency alone does not help
($0.483$); the two together reach $1.000$. Consistent with the illusion critique, the
single-call agent under both levers ($0.983$) matches the panel to within noise at
roughly one-third the calls, so the multi-role decomposition does \emph{not} clearly
earn its extra cost on this task. The ablation's common seed range (base $20{,}000$,
disjoint from both train and the $n{=}200$ eval) is by chance harder than the eval set,
a larger share of confusable-thermal-fault draws ($53\%$ vs.\ $47\%$) at lower mean
severity, which is why the zero-shot panel scores $0.500$ here versus $0.685$ on the
eval seeds; the \emph{relative} lever comparison, run on identical seeds, is unaffected.
How sensitive the lift is to demo \emph{count} and to curated-vs-random \emph{selection}
is the natural follow-up; the released \texttt{scripts/run\_demo\_sensitivity.py} sweeps
$k\in\{0,1,2,3,6,12\}$ and curated-vs-random selection under the identical protocol, a live
LLM run we leave to future work.

\begin{table}[t]
\caption{Ablation on $n{=}60$ common held-out seeds (disjoint from train, eval, and
demos). Classification accuracy by lever and by architecture. Both levers together
drive the panel from $0.500$ to $1.000$; few-shot is the load-bearing lever
(self-consistency alone does not help), and the single-call agent matches the panel
under identical levers ($0.983$) at a fraction of the calls. Verbatim from
\texttt{ablation\_results.json}.}
\label{tab:ablation}
\centering
\begin{tabular}{lcc}
\toprule
\textbf{Condition} & \textbf{Arch.} & \textbf{Class.\ acc.}\\
\midrule
zero-shot & panel & \ABLpanelZero\\
$+$ few-shot only & panel & \ABLpanelFS\\
$+$ self-consistency only & panel & \ABLpanelSC\\
$+$ both & panel & \ABLpanelBoth\\
zero-shot & single & \ABLsingleZero\\
$+$ both & single & \ABLsingleBoth\\
\bottomrule
\end{tabular}
\end{table}

\begin{figure}[t]
\caption{Ablation. Green bars are the five-role panel across levers; violet bars are the
single-call agent under matched levers. Comparing matched pairs isolates whether the
multi-role decomposition earns its extra calls.}
\label{fig:ablation}
\end{figure}

\section{Continuous Monitoring}\label{sec:monitor}
\begin{figure}[t]
\centering
\includegraphics[width=\columnwidth]{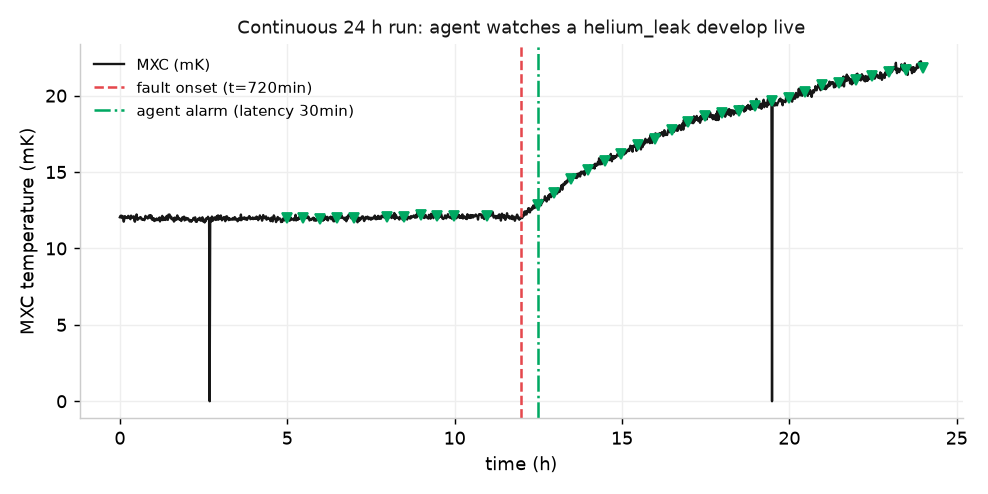}
\caption{Continuous $24$\,h run. MXC rises after the hidden leak onset (red);
the agent's first true alarm (blue) comes $29.5$\,min later. Pre-onset false alarms
are visible as scattered markers.}
\label{fig:monitor}
\end{figure}

The head-to-head scores independent windows, but a real fridge is one unbroken stream:
it runs at base temperature, a fault begins at a hidden moment, and the operator's value
is catching it \emph{early}. We first walk through a \textbf{single illustrative run} to
fix intuition, then generalize it to a $\SWEEPcells$-run fault$\times$seed sweep
so the claim rests on a distribution rather than one trace. For the
illustrative run we simulate a single continuous $24$\,h run at $1$-min
cadence with a helium leak beginning at $t{=}719.5$\,min, and poll the Sentinel agent
every $30$\,min over a rolling look-back window (Fig.~\ref{fig:monitor}).

The agent issues its first true alarm at $t{=}749.0$\,min, a \textbf{detection latency
of $29.5$\,min}, as the MXC climbs from its $\sim\!12$\,mK base. This is a latency, not
a physical lead time: with a $30$\,min poll cadence the smallest observable value is one
poll, so the figure is cadence-bounded and a cadence/seed sweep is future work. Honesty
requires reporting the cost: the Sentinel raised $11$ pre-onset false alarms across $41$
polls, i.e.\ high sensitivity at low precision on the noisy baseline. We stress that this
is a single-seed, single-fault demonstration; a full multi-seed, multi-fault,
variable-cadence study with precision--recall-over-time and latency/false-alarm
trade-off curves is needed before any claim of continuous-monitoring readiness, and is
the most important next experiment. The transferable result here is not the $29.5$\,min
number but the confidence-gating mechanism below, which needs no additional runs.

\textbf{Confidence gating:} The Sentinel already emits a confidence level on every poll,
and the false alarms are mostly low/medium confidence while the true post-onset alarms
are almost all high (Table~\ref{tab:gating}). A post-hoc gate therefore trades precision
for latency cheaply and with no new runs: requiring at least \emph{medium} confidence
halves the false alarms ($11\!\to\!6$) at \emph{no} latency cost, and requiring
\emph{high} confidence nearly eliminates them ($11\!\to\!1$) at the price of one extra
poll of latency ($29.5\!\to\!59.5$\,min). This is a simple, honest fix for the monitor's
precision problem.

\begin{table}[t]
\caption{Post-hoc confidence gating on the $41$ continuous-monitor polls. Numbers
verbatim from \texttt{monitor\_gating.json}. Medium gating removes false alarms for free;
high gating trades one poll of latency for near-zero false alarms.}
\label{tab:gating}
\centering
\resizebox{\columnwidth}{!}{%
\begin{tabular}{lccc}
\toprule
\textbf{Gate} & \textbf{False alarms} & \textbf{Latency (min)} & \textbf{Detected?}\\
\midrule
any (original) & $11$ & $29.5$ & yes\\
medium$+$ & $6$ & $29.5$ & yes\\
high & $\mathbf{1}$ & $59.5$ & yes\\
\bottomrule
\end{tabular}%
}
\end{table}

\begin{figure}[t]
\centering
\includegraphics[width=0.9\columnwidth]{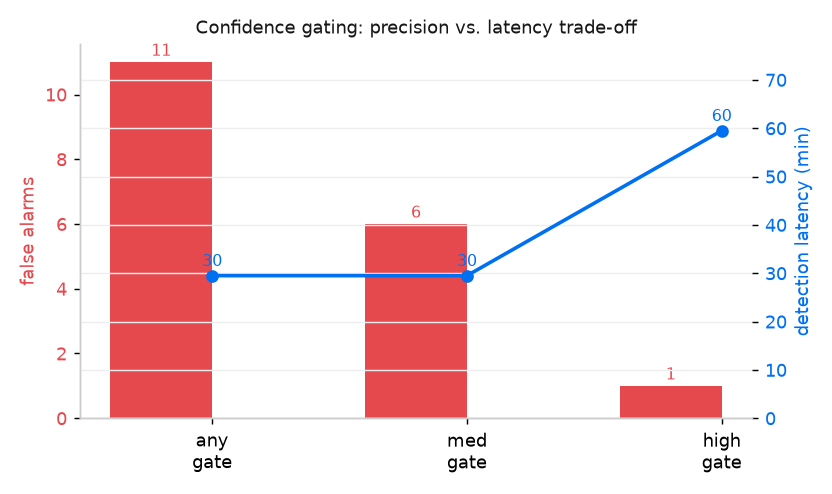}
\caption{Confidence gating trades precision for latency, on the single 24\,h run.
Pre-onset false alarms (red bars) fall as the gate tightens; the medium gate is free, the
high gate costs one poll of detection latency (blue). The multi-seed, multi-fault sweep
below finds this false-alarm behavior is backend-dependent (zero on Gemini~3.1~Pro).}
\label{fig:gating}
\end{figure}

\textbf{From one run to a sweep.} The single run above is an illustrative demonstration,
not evidence of monitoring readiness: one seed, one fault, and a false-alarm count that
could be a property of that one noise realization. To test whether its core claim ---
detection within roughly one poll of onset --- survives across faults and noise
realizations, we ran a live sweep of the identical Sentinel monitor over a
$\SWEEPfaults\times\SWEEPseeds$ grid (\SWEEPfaults\ fault classes spanning slow, sharp,
and step dynamics $\times$ \SWEEPseeds\ seeds $=\SWEEPcells$ continuous 24\,h runs,
$\SWEEPpolls$ live polls each), on the independent Gemini~3.1~Pro backend. Nothing is
re-implemented: each cell is one \texttt{continuous\_monitor.run()} and the gate
re-scoring reuses the same confidence ranking, so every number still traces to a logged
poll from a real agent turn.

Across the grid the agent detects the developing fault in every cell (detection rate
$\SWEEPdetRate$, $\SWEEPcells/\SWEEPcells$), at an overall median latency of
$\SWEEPlatMed$\,min: $\SWEEPfloorN$ of \SWEEPcells\ cells fire at the one-poll floor and a
single helium-leak seed takes two polls ($\SWEEPlatMax$\,min). The one-poll-of-onset
behavior is therefore the typical case, not a lucky seed --- and it stays cadence-bounded
by the $30$\,min poll interval, so the latency number reflects the schedule, not a
physical lead time. The false-alarm picture is where the sweep is most informative, and
we report it honestly: on Gemini~3.1~Pro \emph{every} cell logged $\SWEEPfa$ pre-onset
false alarms, whereas the single Opus run in Table~\ref{tab:gating} logged
$\SWEEPsingleFA$. Pre-onset false-alarm rate is thus \emph{backend-dependent}, not a fixed
property of the monitor: on the quieter backend the confidence gate is a zero-cost safety
margin (nothing to prune), while on the noisier one it is the load-bearing filter that
cut $11\!\to\!1$. This is exactly the argument for shipping the gate as a tunable knob
rather than a fixed threshold. Two caveats keep this from being overread. First, the
comparison is \emph{not} matched: it is $\SWEEPcells$ Gemini cells against a single Opus
run, so the honest statement is ``zero across nine Gemini cells vs.\ eleven in one Opus
run,'' and a matched Opus sweep is required before attributing the difference to the model
rather than to seed variance --- we mark it as future work. Second, the mechanism is
diagnosable from the turn logs: the Opus false alarms are Sentinel flagging normal
noise excursions on the noisiest channels (50\,K and flow) at low/medium confidence before
onset, which is why the medium gate removes them for free; a more robust Sentinel would
condition on a short baseline window per channel rather than an absolute deviation, and
testing that against post-hoc gating is the natural follow-up. The sweep does not add
real-hardware validation (no labeled
real fault data exists; Section~\ref{sec:discussion}), but it converts the monitoring
section from an $n{=}1$ anecdote into a distribution over faults, seeds, and backends ---
which is what monitoring readiness actually requires. All $\SWEEPcells$ runs are released
verbatim in \texttt{monitor\_sweep\_gemini.json}.

\section{Discussion, Limitations, and Future Work}\label{sec:discussion}
\textbf{What the results support:} (i) On realistic, non-saturated telemetry, a
zero-shot agent panel matches supervised ML on \emph{detection}
(McNemar $p=0.07$, no significant difference) but not \emph{classification}
(McNemar $p<10^{-15}$); (ii) the classification gap is concentrated on physically
confusable faults; (iii) two evidence-backed in-context techniques close that gap in a
controlled comparison, with no training; (iv) all three effects replicate on a second
independent model (Gemini~3.1~Pro), so they are properties of the task rather than of one
LLM (Table~\ref{tab:backend}). The negative results, avoiding debate and
self-refinement, are themselves contributions grounded in 2026
literature~\cite{illusion,selfcorrect}.

\textbf{Statistical reporting:} All accuracies carry Clopper--Pearson exact $95\%$
confidence intervals (Table~\ref{tab:headtohead}), and every head-to-head comparison
uses the exact paired McNemar test on identical scenarios, so ``detection ties'' and
``ML wins classification'' are statistical statements, not point-estimate impressions.

\textbf{Ablation, and what the multi-agent structure is actually for:}
Table~\ref{tab:ablation} decomposes the enhancement: the lift is
almost entirely attributable to the curated contrastive demonstrations (few-shot alone
$0.500\!\to\!0.983$), while self-consistency alone does not help ($0.483$). We then run
the test that most ``multi-agent'' papers omit and that recent work argues is the crux of
the multi-agent debate~\cite{illusion}: the five-role panel against a
single well-prompted LLM call under matched levers. The honest result is that a
single-call agent under both levers ($0.983$) matches the panel to within one scenario at
roughly one-third the calls --- \emph{the multi-role decomposition buys no classification
accuracy}. We report this as a finding, not a footnote: on this task, stacking roles for
accuracy is exactly the compute-for-coordination illusion~\cite{illusion}, and a single
prompt is the right choice if accuracy is all you want.

The panel earns its cost on a different axis --- \emph{auditability and safety}, which is
the axis that matters when an autonomous agent can send a command to real hardware. This
is not a consolation prize; it is a deliberate design principle for safety-critical
operations, and it is separable from accuracy in three concrete ways a single call cannot
replicate. (1) \emph{Independent operating points}: the high-recall Sentinel, the
classifying Diagnostician, and the accountable Supervisor can each be tuned and audited at
a different threshold --- a deliberately over-sensitive Sentinel gated by a conservative
Supervisor gives high recall on developing faults without paying the false-alarm cost at
the final verdict, a trade-off a monolithic call cannot expose because it commits to one
threshold. (2) \emph{A single, inspectable veto surface}: the Guardian is one choke point
at which any action that could quench the magnet or warm the mixing chamber is blocked
\emph{before} it reaches hardware, on a code path independent of the classification that
may itself be wrong --- so a misdiagnosis does not silently become a dangerous command.
(3) \emph{A per-role audit trail}: every role's JSON reply is logged, so a post-incident
review can see \emph{which} role failed (did the Sentinel miss it, or did the Guardian
wave through a bad action?), whereas a single call collapses the whole decision into one
opaque output. A single-call agent that emits the same accuracy offers none of these
surfaces. Whether this structure prevents a real incident is an operational claim we
cannot settle on simulated telemetry, and we flag it as such --- but the separation is
the reason to keep the panel, and it is orthogonal to the accuracy the ablation measures.

\textbf{Cross-backend replication (Claude vs.\ Gemini):} A single-backend result cannot
distinguish a property of \emph{the task} from a quirk of \emph{one model}. We therefore
re-ran the entire $n{=}200$ head-to-head on a second, independently-trained frontier
model, Google Gemini~3.1~Pro (\texttt{gemini-3.1-pro-preview}) via the Google GenAI SDK,
holding everything else fixed: the same seed-addressed scenarios, the same five-role
prompts, the same JSON parser, the same supervised RF opponent, and the same scorer, so
the LLM is the only thing that changes. The paper's core finding replicates
(Table~\ref{tab:backend}). Zero-shot, Gemini~3.1~Pro also shows no significant detection
gap ($F_1=\GEMzeroDet$) and \emph{trails} on classification (\GEMzeroCls, vs.\ Claude's
$0.685$), and its errors fall on the same physically-confusable cluster: all $57$
zero-shot misclassifications are among the four warming faults, led by the same two
pairs (\texttt{wiring\_heat\_ingress}$\rightarrow$\texttt{heat\_load\_spike} and
\texttt{helium\_leak}$\rightarrow$\texttt{blocked\_impedance}) that dominate Claude's
confusion matrix (Fig.~\ref{fig:agentconf}). The two in-context levers then lift Gemini's
classification to \GEMenhCls\ on the same seeds, the same qualitative recovery seen for
Claude. That two models from different labs land within $\sim\!0.03$ of each other and
fail on the \emph{same} physics-defined faults is evidence that ``zero-shot LLMs match
supervised detection, lose on classification precisely where the physics is ambiguous,
and recover with curated in-context examples'' is a property of the task, not of one
vendor's model.

\begin{table}[t]
\caption{Cross-backend replication on identical $n{=}200$ seeds ($\mathrm{sev}\times0.5$),
same prompts/parser/opponent; only the LLM differs. Claude via an OpenAI-compatible proxy,
Gemini via the Google GenAI SDK. Verbatim from \texttt{gemini\_vs\_claude.json} and the
Claude \texttt{agent\_eval} artifacts.}
\label{tab:backend}
\centering
\resizebox{\columnwidth}{!}{%
\begin{tabular}{lcc}
\toprule
\textbf{Backend (zero-shot)} & \textbf{Detection $F_1$} & \textbf{Classification acc.}\\
\midrule
Claude Opus~4.8 & $0.979$ & $0.685$\\
Gemini~3.1~Pro & $\GEMzeroDet$ & $\GEMzeroCls$\\
\midrule
\textbf{+ few-shot + self-consistency} & & \\
Claude Opus~4.8 & $0.994$ & $0.990$\\
Gemini~3.1~Pro & $1.000$ & $\GEMenhCls$\\
\midrule
Supervised RF (shared) & $0.997$ & $0.985$\\
\bottomrule
\end{tabular}%
}
\end{table}

\textbf{Limitations:} The enhanced-panel controlled comparison (Table~\ref{tab:lift})
is an $n{=}24$ held-out proof-of-concept: the $24$ scenarios are simply the held-out
draws on which the zero-shot panel scored $0.50$ (all errors confusable-pair
confusions), \emph{not} a subset selected to favor the enhancement; the levers were
fixed in advance. A perfect score at that size cannot by itself support ``matches
supervised ML'' at scale, so we also report the enhanced panel at $n{=}200$
(Table~\ref{tab:lift}) on the same seeds as the zero-shot anchor. \emph{Two threats to the
enhanced result deserve explicit statement.} (i)~\textbf{No held-out dev split}: the levers
($k{=}6$, $N{=}3$, $\tau{=}0.7$) and the confusable-pair demonstration weighting are fixed
using knowledge of where the zero-shot panel fails on the evaluation scenarios, not on a
separate tuning set (Table~\ref{tab:hyper}). The $0.990$ is therefore best read as a
configuration tuned toward this eval, i.e.\ an optimistic upper bound; the cross-backend
replication on Gemini ($0.715\!\to\!0.995$, Table~\ref{tab:backend}) shows the effect
transfers across models but does not remove the concern, and a blind dev/eval split is the
clean fix. (ii)~\textbf{Parity is parity on a near-saturated task}: both the enhanced panel
($0.990$) and supervised ML ($0.985$--$0.996$) sit near ceiling because the twin's fault
signatures are authored at high SNR (Table~\ref{tab:baselines}); a perfect-ish score on a
benchmark designed to be separable is weak evidence for real-hardware parity. The
sensor-noise stress test (Fig.~\ref{fig:stress}, RF holds $\ENGnoiseRF$ macro-$F_1$ at
$20\%$ noise) indicates the harder, non-saturated regime is where the comparison is most
informative, and hardening the benchmark until neither system saturates is future work. The
continuous-monitoring claim now rests on a $\SWEEPcells$-run sweep
(Section~\ref{sec:monitor}) rather than one run, but detection
latency there remains bounded below by the $30$\,min poll cadence, so the
$\SWEEPlatMed$\,min figure is a detection \emph{latency} set by the schedule, not a
physical lead time, and the sweep is still simulated telemetry. The twin, while
physics-grounded and fingerprinted to real logs, is still a simulator (a forward model,
not a hardware-coupled bidirectional twin); validation against labeled real-fridge fault
episodes is the ultimate test. We also state the fidelity gap quantitatively rather than
rhetorically: a random-forest classifier two-sample test~\cite{lopezpaz2017} trained to
tell twin windows from real BlueFors windows achieves cross-validated
$\mathrm{AUC}=1.00$ (permutation null $0.53$; $42$ real vs.\ $42$ twin $3$\,h windows), in
both absolute and level-invariant ``fingerprint'' feature conditions
(\texttt{onnesim.twin\_fidelity}, reproducible in $\sim\!2$\,min). The fingerprint matches
the real per-stage noise magnitudes and cross-stage correlations to within a few percent
--- close enough that the fault \emph{contrasts} are realistic --- but a discriminative
model still separates the marginal distributions perfectly. The twin is therefore a
controlled benchmark environment, \emph{not} a statistically indistinguishable replica of
any specific fridge, and every accuracy in this paper should be read as ``on the twin.''
The feature importances of that discriminator name exactly which channels to correct
first, which is the concrete entry point for stage~(1) of the plan below.

\textbf{Closing the sim-to-real gap --- a concrete validation plan.} Because that gap is
the load-bearing caveat for every deployment claim in this paper, we state the validation
path explicitly rather than gesturing at ``future work,'' in three stages of increasing
cost and evidential weight. \emph{(1) Replay on real healthy telemetry --- done, first
result.} We take the first stage off the roadmap and run it: an anomaly detector is trained
\emph{only} on real BlueFors healthy windows ($281$ windows, no twin, no synthetic normal;
a PCA reconstruction detector, the field-standard approach under labeled-fault scarcity) and
its false-alarm rate is measured on $188$ held-out \emph{real} healthy windows. The result
is a genuine real-hardware number: $6.4\%$ pre-onset false alarms on real nominal telemetry
(\texttt{onnesim.real\_detect}, deterministic, reproducible in ${<}1$\,min). On the same real
held-out windows, physics-grounded faults injected onto the real noise floor (the accepted
``physics-guided augmentation'' protocol under fault-label scarcity, applied to a dilution
fridge for the first time) are caught at $100\%$ recall across all five injectable classes.
This is strictly stronger evidence than any twin-only number --- the normal distribution and
noise are entirely real --- though it still cannot claim real fault \emph{classification},
which needs real labels (stage~2). It confirms the monitor's low false-alarm behavior
survives real drift, the cheapest deployment check and the one that needs no fault to occur.
\emph{(2)
Retrospective labeled episodes (months).} Assemble a corpus of historical fault episodes
(quench, helium leak, blocked impedance) from operator logbooks and maintenance tickets,
each a diagnosed window with a start time; replay them to measure real detection latency
and classification accuracy against the $6$-class taxonomy. Roughly $5$--$10$ episodes per
class is enough to move the classification claim from ``matches supervised ML on the twin''
to ``matches supervised ML on real faults,'' and is the single most valuable artifact a
partner lab could contribute. \emph{(3) Prospective shadow deployment (months, live).} Run
the monitor read-only alongside a commissioned fridge, gating alarms to an operator
dashboard (never to control), and score latency and false-alarm rate on faults as they
naturally occur --- the only test that exercises the $30$\,min cadence against real onset
dynamics. Stage~(1) is now done (above); stage~(2) needs no privileged hardware access,
only labeled log archives, and is the immediate next step; stage~(3) is the eventual bar
for an operational monitor.

\textbf{Stronger opponents:} We evaluate a baseline zoo (random forest, histogram
gradient boosting, LightGBM, and the TabPFN-2.5 tabular foundation
model~\cite{tabpfn25}) under one seed-addressed protocol (Table~\ref{tab:baselines}).
TabPFN-2.5 is the strongest supervised opponent (macro-$F_1\;0.996$), slightly ahead of
the random forest ($0.983$); we keep the random forest as the primary fully-local
opponent and report TabPFN-2.5 as the stronger foundation-model bar. That the enhanced
training-free panel ($0.990$) reaches this ceiling is the central result, and we expect
the qualitative story (detection ties, classification hinges on data/priors) to
persist against even stronger tabular models.

\textbf{Cost, and what actually differs:} The panel is $\sim\!5$ LLM calls per window at a
measured $\sim\!17$\,s each ($7$ calls with $N{=}3$ voting), versus a random forest that
predicts a window in $<\!1$\,ms after a sub-second fit (Table~\ref{tab:cost}). Speed is
not the agent's advantage. We are also careful not to overstate the label argument: in
\emph{this} study the RF's $300$ training labels and the agent's $6$ demonstrations are
\emph{both} free simulator output, so the study does not by itself show a labeling-cost
saving. What it does show, on identical scenarios, is a $6$-vs-$300$ example-count gap
(the enhanced panel reaches supervised accuracy from $6$ demonstrations and no parameter
updates) together with interpretable, physics-grounded reasoning and a per-turn audit
trail the RF does not provide. The \emph{economic} case is therefore conditional and
belongs to deployment, not to this simulator: it holds only insofar as the twin transfers
to real hardware, because on a real fridge a label is not free metadata but a
\emph{diagnosed fault episode} (expert operator time to identify and annotate a real
leak, quench, or blockage, and, for rare classes, waiting for one to occur). If the
twin-to-reality gap is closed (Section~\ref{sec:discussion}, ``Limitations''), a method
that needs $6$ curated demonstrations rather than $300$ diagnosed episodes is the cheaper
route to a first diagnosis on a newly-commissioned fridge with no fault history; the RF
becomes preferable once a labeled corpus already exists. We state this as a conditional,
not a demonstrated result.

\begin{figure*}[t]
\centering
\includegraphics[width=\textwidth]{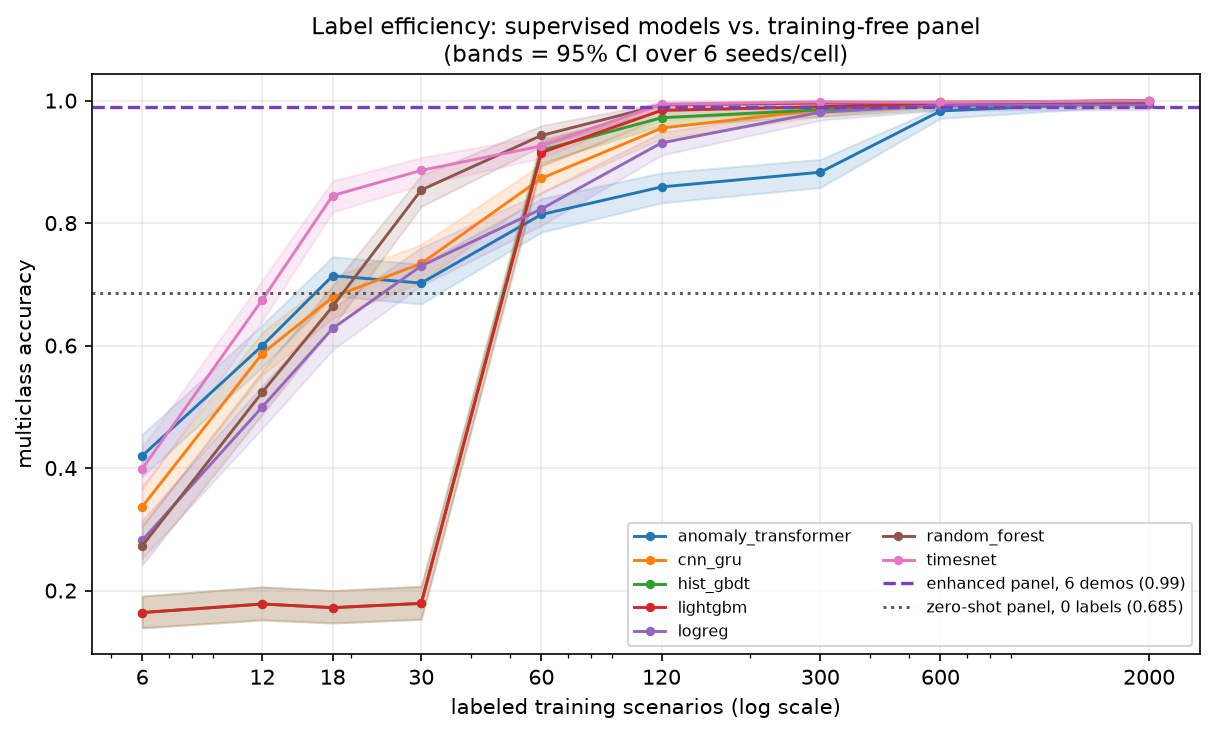}
\caption{Label efficiency: multiclass accuracy ($y$) versus number of labeled training
scenarios ($x$, log scale) on identical held-out seeds (severity fixed; $6$ seeds/cell).
Shaded bands are $95\%$ CIs over the $6$ trainings per cell; the dashed violet line is the
enhanced panel's six-demonstration accuracy ($0.990$) and the dotted grey line the
zero-shot panel ($0.685$), both training-free references (flat because they use no training
labels). Supervised opponents (tabular:
random forest, hist-GBDT, LightGBM, logistic; deep: CNN-GRU, TimesNet~\cite{timesnet}, Anomaly-Transformer~\cite{anomalytransformer})
need $120$--$300$ labeled episodes to reach the enhanced panel; at six labels they score
$0.16$--$0.42$. All values are on the twin.}
\label{fig:labeff}
\end{figure*}

\textbf{Label efficiency (quantifying the $6$-vs-$300$ gap):} To make the example-count
argument concrete rather than assert it, we sweep the \emph{training-label budget} for every
supervised opponent at fixed severity, evaluating on the identical held-out seeds the agent
faces, with $6$ repeated trainings per cell (Fig.~\ref{fig:labeff}, verbatim from
\texttt{label\_efficiency.json}). The tabular and deep zoos alike are near-perfect once given
several hundred labels, but collapse in the few-label regime a real fridge actually occupies:
at \emph{six} labels the strongest deep opponent reaches only $0.42$ and the gradient-boosted
models sit at chance ($0.16$), whereas the enhanced panel classifies at $0.990$ from six
demonstrations. Supervised parity with that six-demonstration accuracy is not reached until
roughly $120$--$300$ labels ($20$--$50\times$ more), and only then does the ML advantage of
Table~\ref{tab:headtohead} reappear. This is the honest shape of the label argument: not that
the agent is more accurate at scale (it is not), but that it reaches supervised-level accuracy
at a label budget where every supervised model we tested is far below it. As with every other
number here, this is measured on the twin; the relative comparison (agent vs.\ ML at equal
label budget, identical seeds) is what the figure claims, not a real-hardware result.

\subsection{Deployment considerations}\label{sec:deployment}
Four practical constraints shape how this would run
against real fridges, and we flag them so the operational-value claim is not read as
turnkey. \textbf{(i) Multi-agent premium and concurrency.} The panel is $5$--$7$ LLM calls
per window; our ablation (Table~\ref{tab:ablation}) shows the multi-role decomposition
buys auditability and a safety-veto seam, \emph{not} accuracy --- a single call with the
same demonstrations matches it at roughly one-third the calls. A fleet of $N$ fridges
polled concurrently should therefore default to the single-call diagnostician for cost,
escalating to the full panel only when the Sentinel gate fires, keeping the panel premium
where its audit trail earns it. \textbf{(ii) Guardian robustness (deterministic interlocks).}
The Guardian is currently
an LLM policing an LLM; under a genuinely out-of-distribution ``black-swan'' fault, the
diagnostic and safety roles can share a blind spot. The robust production form binds the
Guardian to deterministic programmatic interlocks (magnet-current, MXC warm-rate, and
heater ceilings read straight from telemetry), using the LLM only to \emph{explain} a veto
the hardware limit already enforces. \textbf{(iii) Serialization and cadence/token
trade-off.} We present the
window as a compact per-channel summary rather than raw rows, which bounds token cost at
the $5$\,min cadence studied here; catching sub-second transients (fast valve chatter,
micro-oscillations) would need either a higher-rate summary or an event-triggered
hand-off, and the token/latency trade-off of that regime is unmeasured.
\textbf{(iv) Static demonstrations and long-term drift.} The enhanced result rests on a fixed
$k{=}6$ contrastive demonstration block. On a real fridge subject to months-long drift
(lab-temperature swings, pump wear, gas-mixture degradation), a static block can go stale;
the released retrieval path (\texttt{retrieval.py}, which selects demonstrations by nearest
neighbor in feature space) is the natural hook for a periodically-refreshed demonstration
pool, but we have not evaluated drift adaptation and treat it as open.

\textbf{Future work:} (1) further LLM backends beyond the Claude/Gemini replication
reported here, and finer-grained monitoring sweeps over poll cadence (the
$\SWEEPcells$-run fault$\times$seed sweep of Section~\ref{sec:monitor} fixes the cadence at
$30$\,min); (2) a TabPFN-2.5 column under the released protocol; (3) stages~(2)--(3) of the
real-fridge validation plan above (retrospective labeled episodes $\to$
prospective shadow deployment; stage~(1), real healthy-log replay, is done), which is the
decisive next step; (4) query-conditioned
retrieval for demonstration selection at larger demo banks, where a fixed round-robin block
no longer saturates; (5) a deterministic-interlock Guardian and a study of its veto
behavior under injected out-of-distribution faults; (6) transfer of the twin-design
methodology (physics mean $+$ learned noise fingerprint $+$ deliberately confusable fault
signatures) to other cryogenic instruments and, more broadly, to scientific-instrument and
industrial-control settings where the same label-scarcity and confusable-fault structure
recurs; and (7) integration with existing fridge operations (BlueFors control software,
Grafana/Slack alarm pipelines, and operator escalation), for which the released
\texttt{make\_dashboard.py} is a first proof-of-concept surface.

\section{Reproducibility}\label{sec:repro}
Every number in this paper is emitted by a run script and read back from a JSON/JSONL
artifact; none are typed by hand. Table~\ref{tab:repro} maps each claim to its source.
The evaluation harness logs all $1000$ agent turns; the head-to-head reconstruction is
computed directly from that log, so a run interrupted mid-way still yields an honest
scorecard over completed scenarios. Seed ranges for ML training ($0$--), few-shot demos
($500$--), and evaluation ($10{,}000$--) are disjoint by construction and unit-tested.
Agent turns are Claude Opus~4.8 queried through an OpenAI-compatible proxy at a pinned
model id; the cross-backend replication uses Gemini~3.1~Pro
(\texttt{gemini-3.1-pro-preview}) through the Google GenAI SDK
(\texttt{scripts/eval\_gemini.py}). The exact ids and the code are released with the
artifacts.\footnote{Code and run
logs: \url{https://github.com/Onnes-Research/onnes}.} Statistical tests (Clopper--Pearson
CIs, exact McNemar) are recomputed from the released logs by \texttt{onnesim/stats.py},
and the confidence-gating and cost analyses by \texttt{monitor\_gating.py} and
\texttt{cost\_model.py}.

\textbf{Compute.} The tabular baselines and the agent evaluation run on CPU; the deep
opponents and the label-efficiency sweep were trained on a single NVIDIA~B200 (180\,GB)
via \texttt{scripts/run\_label\_efficiency.py}. The label-efficiency grid
(Fig.~\ref{fig:labeff}: $7$ models $\times$ $9$ label budgets $\times$ $6$ seeds, $378$
trainings, evaluated on a fixed $800$-scenario held-out set, deep models at $250$ epochs
with early stopping) completed in $250$\,s of wall-clock; the single-severity deep zoo
($3$ architectures $\times$ $8$ seeds on $6000$ training windows) in $639$\,s. GPU memory is
not the constraint---these models are small---so the hardware buys \emph{breadth} (many
seeds and budgets for tight confidence intervals), not model size; the same scripts
reproduce the numbers on any CUDA or Apple-MPS device via the \texttt{ONNES\_DEVICE}
environment variable.

\begin{table}[t]
\caption{Claim-to-artifact map. All results are regenerated from these files.}
\label{tab:repro}
\centering
\footnotesize
\begin{tabular}{ll}
\toprule
\textbf{Claim} & \textbf{Source artifact}\\
\midrule
Zero-shot $0.979$/$0.685$ ($n{=}200$) & \texttt{agent\_eval\_results.json}\\
Head-to-head CIs $+$ McNemar & \texttt{head\_to\_head\_stats.json}\\
Enhanced lift ($n{=}24$) & \texttt{technique\_lift.json}\\
Enhanced panel ($n{=}200$) & \texttt{agent\_eval\_fewshot\_n200\_}\\
 & \texttt{results.json}\\
Ablation (levers $+$ arch.) & \texttt{ablation\_results.json}\\
Baseline zoo & \texttt{baseline\_zoo.json}\\
Cost / latency & \texttt{cost\_model.json}\\
Monitor $29.5$\,min latency, gating & \texttt{continuous\_monitor.json},\\
 & \texttt{monitor\_gating.json}\\
Monitor sweep ($\SWEEPcells$ runs, Gemini) & \texttt{monitor\_sweep\_gemini.json}\\
ML stress macro-$F_1$ & \texttt{benchmark\_results.json}\\
Label-efficiency curve (Fig.~\ref{fig:labeff}) & \texttt{label\_efficiency.json}\\
All figures & \texttt{scripts/make\_figures.py}\\
\bottomrule
\end{tabular}
\end{table}

\section{Conclusion}
Onnes couples a physics-grounded, fingerprinted dilution-refrigerator twin to a live
multi-agent LLM operations layer and uses it for a controlled study of agents versus
supervised ML on cryogenic fault diagnosis. The honest headline is not that agents win:
zero-shot, they show \emph{no significant difference} on detection and \emph{lose} on
classification, failing
exactly on the faults engineered to be confusable. The constructive result is that two
carefully-chosen, evidence-backed in-context techniques (contrastive few-shot and
self-consistency, and pointedly \emph{not} debate or self-refinement) close that gap in
a controlled comparison with no task-specific training. For infrastructure where faults
are rare and labels scarce, a training-free method that reaches supervised accuracy is
operationally significant, and the twin makes such claims measurable and reproducible.

This work was carried out at \textbf{Onnes Research}, the research group of
\textbf{Onnes} (\href{https://onnes.ai}{onnes.ai}); the digital twin, agent layer, and
all released run logs are open-source at \href{https://github.com/Onnes-Research/onnes}%
{github.com/Onnes-Research/onnes} to support reproducible research on cryogenic fault
diagnosis for quantum-computing infrastructure.

\appendix
\section{Panel and Auxiliary Prompts}\label{app:prompt}
\sloppy
The five panel roles (A.1--A.5) are single LLM calls with the system prompts below,
verbatim from \texttt{onnesim/\allowbreak multi\_agent.py} (\texttt{run\_panel}). Each role replies in
strict JSON; the window \texttt{\{summary\}} is the per-channel start/end/\%-change plus
coarse trajectory produced by \texttt{summarize\_window}. Only the Diagnostician's
\emph{user} message changes between conditions (the optional demonstration block); every
system prompt is fixed. The prompts are deliberately terse: the paper's claim is that the
contrastive demonstrations, not prompt engineering, drive the accuracy gain
(Section~\ref{sec:incontext}), so hand-tuning the base prompts would confound that
ablation. Beyond the five panel roles, the released code contains two auxiliary prompts
that other reported numbers depend on --- the single-call baseline of Table~\ref{tab:cost}
(``Single agent'') and a selective Verifier --- reproduced verbatim in A.6--A.7 for
completeness.

\smallskip
\noindent\textbf{A.1\quad Sentinel.}\\
{\footnotesize\ttfamily System: You are Sentinel, watching dilution-fridge telemetry.
Reply ONLY JSON: \{"anomaly\_developing": bool, "which\_channels": [str], "confidence":
"low|med|high"\}.\\
User: Telemetry window summary:\{summary\} Is an anomaly developing?}

\smallskip
\noindent\textbf{A.2\quad Diagnostician.}
The Diagnostician's system prompt is identical across conditions; only the
\emph{user} message differs by an optional contrastive-demonstration block:

\smallskip
\noindent\textbf{System (all conditions):}
{\footnotesize\ttfamily
You are Diagnostician for a Bluefors dilution fridge. Stages temp1=50K,
temp2=4K(magnet flange, quench risk), temp3=still, temp4=cold plate,
temp5=mixing chamber(\textasciitilde{}10mK), temp6/7=magnet. Fault classes: normal,
heat\_load\_spike, helium\_leak, magnet\_quench, wiring\_heat\_ingress,
blocked\_impedance. Reply ONLY JSON: \{"fault\_class": str, "severity":
"none|low|medium|high", "reason": str\}.
}

\smallskip
\noindent\textbf{User (zero-shot):} \texttt{\footnotesize Telemetry:\{summary\}\textbackslash nClassify.}

\smallskip
\noindent\textbf{User (few-shot):} the same, prefixed with $k{=}6$ labeled
\texttt{\{summary\}$\rightarrow$fault\_class} demonstrations that over-weight the
confusable pairs, drawn from a seed range disjoint from train and eval. Under
self-consistency the user message is unchanged; the call is sampled $N{=}3$ times at a
diversity temperature and the modal \texttt{fault\_class} is taken.

\smallskip
\noindent\textbf{A.3\quad Operator.}\\
{\footnotesize\ttfamily System: You are Operator. Propose ONE corrective action for the
diagnosed fault. Reply ONLY JSON: \{"action": str, "urgency": "low|medium|high"\}.\\
User: Telemetry:\{summary\} Diagnosis:\{diagnostician\} Action?}

\smallskip
\noindent\textbf{A.4\quad Guardian.}\\
{\footnotesize\ttfamily System: You are Guardian, enforcing fridge safety. Block actions
that could quench the magnet, warm the mixing chamber uncontrollably, or exceed heater
limits. Reply ONLY JSON: \{"approved": bool, "why": str\}.\\
User: Proposed action:\{operator\} Approve?}

\smallskip
\noindent\textbf{A.5\quad Supervisor.}\\
{\footnotesize\ttfamily System: You are Supervisor. Reconcile the panel into a final
verdict. Reply ONLY JSON: \{"fault\_detected": bool, "fault\_class": str, "final\_action":
str, "confidence": str\}.\\
User: Sentinel:\{$\cdots$\} Diagnostician:\{$\cdots$\} Operator:\{$\cdots$\}
Guardian:\{$\cdots$\} Final verdict?}

\smallskip
\noindent The Guardian is presently an LLM-level safeguard, not a hardware interlock;
Section~\ref{sec:discussion} notes that binding it to deterministic programmatic limits
(magnet current, MXC warm-rate, heater ceilings) is the robust production form.

\smallskip
\noindent\textbf{A.6\quad Single-agent baseline.}
The panel-vs-single ablation (Table~\ref{tab:cost}, ``Single agent'') collapses the
five roles into one call that does detection and classification directly, so the
$5\times$ call count is the only variable. The few-shot and self-consistency levers apply
unchanged.\\
{\footnotesize\ttfamily System: You are a dilution-fridge fault diagnostician. Stages
temp1=50K, temp2=4K (magnet flange), temp3=still, temp4=cold plate,
temp5=mixing chamber(\textasciitilde{}10mK), temp6/7=magnet. Fault classes: normal,
heat\_load\_spike, helium\_leak, magnet\_quench, wiring\_heat\_ingress,
blocked\_impedance. In ONE step, decide if a fault is developing and which class. Reply
ONLY JSON: \{"fault\_detected": bool, "fault\_class": str, "confidence": str\}.\\
User: Telemetry window summary:\{summary\} Diagnose.}

\smallskip
\noindent\textbf{A.7\quad Selective Verifier.}
An optional, training-free self-verification pass (\texttt{run\_verifier}): if the panel's
normalized verdict is \texttt{normal}, a skeptic re-reads the raw window and may overturn it
to a fault class, gated to fire \emph{only} on \texttt{normal} verdicts so it is cheap and
biased toward recovering missed faults rather than manufacturing them. It is released but
\textbf{off by default} (\texttt{verify\_on\_normal=False}); the headline enhanced-panel
result ($0.990$, $k{=}6$ demos, $N{=}3$ vote) does \emph{not} invoke it. Reported here for
completeness.

\smallskip

{\footnotesize\ttfamily\sloppy
\emergencystretch=8em
\spaceskip=0.5em plus 0.5em minus 0.1em
\def\_{\char95\allowbreak}
System: You are Verifier, a skeptical second reader for a Bluefors dilution fridge.
The panel concluded this window is NORMAL. Faint faults near the noise floor are exactly
what a panel misses, so re-examine before agreeing. Stage map:
temp3=still, temp4=cold plate, temp5=mixing chamber (\textasciitilde{}10 mK),
flowmeter/p2/p5=flow \& pressures. PHYSICS of the faults a 'normal' verdict can hide:
(-)~wiring\_heat\_ingress: a parasitic CONDUCTED load --- cold plate (temp4) AND mixing
chamber (temp5) BOTH drift UP together (even tens of mK), while still, flow, and all
pressures stay FLAT. Correlated temp4+temp5 rise with flat flow/pressure = this fault,
NOT normal (normal keeps temp4 and temp5 flat). (-)~heat\_load\_spike: mixing chamber
(temp5) rises but cold plate (temp4) stays flat (MXC-only). If temp4 also rises, it is
wiring\_heat\_ingress, not heat\_load\_spike. (-)~helium\_leak / blocked\_impedance:
warm cold stages BUT flow drops and/or pressures move; if flow and pressures are flat it
is not these. Be conservative: only overturn NORMAL when a specific physical pattern is
present. If temperatures, flow, and pressures are genuinely flat, agree it is normal.
Reply ONLY JSON:
\{"is\_fault": bool, "fault\_class": str, "reason": str\}.
Use fault\_class from the six classes above; use "normal" if truly normal.

User: Telemetry window the panel called NORMAL:\{summary\}
Re-examine for a faint fault hiding under a normal verdict.
\par}
\fussy
\section{Algorithms, Feature Vector, and Hyperparameters}\label{app:algs}
\sloppy
For text-only reconstructibility we give the three core procedures in pseudocode
(verbatim to \texttt{cryo\_\allowbreak engine}, \texttt{multi\_\allowbreak agent}, and \texttt{agent\_\allowbreak eval}),
enumerate the $120$-d feature vector, and consolidate every fixed hyperparameter.
\fussy

\begin{algorithm}[H]
\caption{Twin scenario generation \& fault injection}\label{alg:twin}
\begin{algorithmic}[1]
\STATE \textbf{Input:} fault class $c$, severity $s$, onset fraction $o$, seed $\sigma$
\STATE $\bar{\mathbf{T}}(t) \gets$ physics mean via $T^2$ floor (Eq.~\ref{eq:cool}); upper stages pulse-tube fixed
\STATE inject $c$ as heat-load perturbation on $\{$Still,\,CP,\,MXC$\}$ scaled by $s{\times}0.5$ (realism stressor), active after onset $o$
\STATE apply flow/pressure signature of $c$ (leak: flow$\downarrow$, $p_5\uparrow$; block: flow$\downarrow$, $p_1\uparrow$; quench: $\Delta$-of-exponentials pulse)
\STATE $\boldsymbol{\varepsilon}(t)\sim\mathcal{N}(\mathbf 0,\boldsymbol\Sigma_\sigma)$ from BlueFors fingerprint; $\mathbf x(t)\gets\bar{\mathbf T}(t)(1+\boldsymbol\varepsilon)$
\STATE add sensor imperfections (noise floor, dropouts, railing) with seed $\sigma{+}9973$
\STATE \textbf{return} FRTMS-schema window (seed-addressed: $\sigma$ determines every draw)
\end{algorithmic}
\end{algorithm}

\begin{algorithm}[H]
\caption{Panel execution (zero-shot / enhanced)}\label{alg:panel}
\begin{algorithmic}[1]
\STATE $s \gets$ \texttt{summarize\_window}(cols); $B \gets$ few-shot block (or $\varnothing$)
\STATE $r_{\text{sen}} \gets$ Sentinel$(s)$
\IF{$N{=}1$}
  \STATE $r_{\text{dia}} \gets$ Diagnostician$(B\!\parallel\!s)$
\ELSE
  \STATE sample $N$ diagnoses at $\tau{=}0.7$; $r_{\text{dia}} \gets$ modal-\texttt{fault\_class} sample
\ENDIF
\STATE $r_{\text{op}} \gets$ Operator$(s, r_{\text{dia}})$; $r_{\text{gu}} \gets$ Guardian$(r_{\text{op}})$ \COMMENT{safety veto}
\STATE $r_{\text{sup}} \gets$ Supervisor$(r_{\text{sen}},r_{\text{dia}},r_{\text{op}},r_{\text{gu}})$
\STATE each role: \texttt{extract\_json}; on parse-fail return \texttt{\{\_error\}} (no invented verdict); log turn
\STATE \textbf{return} \{sentinel,\dots,supervisor\}
\end{algorithmic}
\end{algorithm}

\begin{algorithm}[H]
\caption{Evaluation harness \& paired stats}\label{alg:eval}
\begin{algorithmic}[1]
\STATE train RF on \texttt{sample\_specs}$(300,\text{base}{=}0)$ \COMMENT{seeds $0$--$299$}
\STATE eval specs $\gets$ \texttt{sample\_specs}$(200,\text{base}{=}10000)$ \COMMENT{disjoint from train + demos}
\FORALL{spec in eval specs (parallel)}
  \STATE cols $\gets$ simulate(spec); agent \& RF both predict on \emph{this same} cols
\ENDFOR
\STATE detection/classification via \texttt{evaluate.score}; Clopper--Pearson $95\%$ CIs
\STATE McNemar: exact binomial on discordant pairs $(b,c)$ over identical scenarios (\texttt{stats.py})
\end{algorithmic}
\end{algorithm}

\smallskip
\noindent\textbf{Feature vector ($120$-d).} The supervised opponent maps each window to
$15$ channels $\times$ $8$ per-channel statistics. Channels:
\texttt{temp1\_T}\dots\texttt{temp8\_T}, \texttt{flowmeter}, \texttt{p1}\dots\texttt{p6}
(the constant \texttt{cpa\_status}/\texttt{turbo\_status} step signals are excluded).
Statistics per channel: \{start, end, min, max, mean, slope (least-squares over
$t\in[0,1]$), pct\_change, std\}. No quantiles or spectral features are used; the derived
\texttt{dQdt\_4K} channel is excluded from features.

\begin{table}[t]
\caption{All fixed hyperparameters, verbatim from source. \emph{These are not tuned on a
held-out dev split}; Section~\ref{sec:discussion} discusses the resulting optimism.}
\label{tab:hyper}
\centering
\resizebox{\columnwidth}{!}{%
\begin{tabular}{lll}
\toprule
\textbf{Component} & \textbf{Parameter} & \textbf{Value}\\
\midrule
LLM (litellm/Claude) & \texttt{max\_tokens} & $4096$\\
LLM (litellm/Claude) & \texttt{temperature},\,\texttt{top\_p} & unset (backend default)\\
LLM (Gemini) & \texttt{max\_output\_tokens} & $8192$\\
LLM (Gemini) & \texttt{thinking\_budget} & $512$\\
Self-consistency & sample count $N$ & $3$\\
Self-consistency & diversity temp.\ $\tau$ & $0.7$\\
Few-shot & demos $k$ & $6$ (confusable-weighted)\\
Few-shot & demo seed range & $500$--$505$ (disjoint)\\
Random forest & \texttt{n\_estimators} & $300$\\
Realism stressor & \texttt{sev\_scale} & $0.5$\\
ML opponent & train / eval $n$ & $300$ / $200$\\
Monitor & poll cadence & $30$\,min\\
Monitor & rolling window & $4$\,h look-back\\
Monitor & run length / $\Delta t$ & $24$\,h / $1$\,min\\
\bottomrule
\end{tabular}%
}
\end{table}

\FloatBarrier   



\begin{thebibliography}{00}
\bibitem{manyshotcot} T.\ T.\ Chung, L.\ Liu, M.\ Yu, and D.-Y.\ Yeung, ``Many-Shot
CoT-ICL: Making In-Context Learning Truly Learn,'' \emph{arXiv:2605.13511}, ICML 2026.
\bibitem{timera} Y.\ Yang, Z.\ Liu, L.\ Song, K.\ Ying, Z.\ Wang, T.\ Bamford,
S.\ Vyetrenko, J.\ Bian, and Q.\ Wen, ``Time-RA: Towards Time Series Reasoning for Anomaly
Diagnosis with LLM Feedback,'' \emph{arXiv:2507.15066}, ACL 2026 Findings.
\bibitem{selfconsistency} X.\ Wang, J.\ Wei, D.\ Schuurmans, Q.\ Le, E.\ Chi, S.\ Narang,
A.\ Chowdhery, and D.\ Zhou, ``Self-Consistency Improves Chain of Thought Reasoning in
Language Models,'' \emph{arXiv:2203.11171}, ICLR 2023.
\bibitem{selfcorrect} A.\ Liu and J.\ Meng, ``Self-Correction as Feedback Control: Error
Dynamics, Stability Thresholds, and Prompt Interventions in LLMs,''
\emph{arXiv:2604.22273}, 2026.
\bibitem{madesign} H.\ Zhou, X.\ Wan, R.\ Sun, H.\ Palangi, S.\ Iqbal, I.\ Vuli\'c,
A.\ Korhonen, and S.\ \"O.\ Ar{\i}k, ``Multi-Agent Design: Optimizing Agents with Better
Prompts and Topologies,'' \emph{arXiv:2502.02533}, ICLR 2026.
\bibitem{illusion} P.\ Jwalapuram, H.\ Lin, C.\ Li, F.\ Jiao, S.\ Wang, Y.\ Ming, Z.\ Ke,
C.\ Qin, G.\ Carenini, and S.\ Joty, ``The Illusion of Multi-Agent Advantage,''
\emph{arXiv:2606.13003}, 2026.
\bibitem{delayedverify} I.\ Itkin, ``Delayed Verification Destabilizes Multi-Agent LLM
Belief: Instability Thresholds and Optimal Corrector Placement,''
\emph{arXiv:2606.27409}, 2026.
\bibitem{tabpfn25} L.\ Grinsztajn, K.\ Fl\"oge, O.\ Key, \emph{et al.}, N.\ Hollmann, and
F.\ Hutter (Prior Labs), ``TabPFN-2.5: Advancing the State of the Art in Tabular
Foundation Models,'' \emph{arXiv:2511.08667}, 2025.
\bibitem{tabgap} M.\ J.\ Kim, M.\ Schambach, F.\ Essenberger, A.\ Sres, and J.\ H\"ohne,
``Exploring Differences Between Tabular Enterprise Data and Public Benchmarks,''
\emph{arXiv:2606.30452}, 2026.
\bibitem{wielgosz} M.\ Wielgosz, A.\ Skocze\'n, and M.\ Mertik, ``Using LSTM Recurrent
Neural Networks for Monitoring the LHC Superconducting Magnets,'' \emph{Nucl.\ Instrum.\
Methods A}, vol.\ 867, pp.\ 40--50, 2017.
\bibitem{tilaro} F.\ Tilaro, B.\ Bradu, M.\ Gonzalez-Berges, \emph{et al.}, ``Model
Learning Algorithms for Anomaly Detection in CERN Control Systems,'' in \emph{Proc.\
ICALEPCS}, 2017.
\bibitem{cacace} P.\ Cacace, D.\ R.\ Santos, L.\ Giusti, F.\ Ferrand, \emph{et al.},
``Machine Learning Framework for Anomaly Detection and Maintenance Optimization in
Large-Scale Cryogenic Systems,'' \emph{IOP Conf.\ Ser.}, 2025.
\bibitem{tennant} C.\ Tennant, A.\ Carpenter, T.\ Powers, \emph{et al.},
``Superconducting Radio-Frequency Cavity Fault Classification Using Machine Learning at
Jefferson Laboratory,'' \emph{Phys.\ Rev.\ Accel.\ Beams}, vol.\ 23, 114601, 2020.
\bibitem{cryometrics} L.\ Chen, ``cryometrics: BlueFors dilution-refrigerator log
samples,'' GitHub repository, \texttt{github.com/larchen/cryometrics}. Fridge
``blizzard,'' logs dated 2021-10-08 (accessed 2026).
\bibitem{leeds} University of Leeds Research Data Repository, ``Cryostat temperature
logs: dataset for `Directed delivery of terahertz frequency radiation within a dry
${}^3$He dilution refrigerator','' CC~BY~4.0,
\texttt{archive.researchdata.leeds.ac.uk/1034} (accessed 2026). DR-200 dilution-fridge
cool-down logs, Sep 2020--Mar 2022.
\bibitem{ornlsns} B.\ Maldonado, D.\ Winder, P.\ Ramuhalli, and W.\ Blokland,
``Process and control variables from the SNS's cryogenic moderator system,'' Oak Ridge
National Laboratory [data set], 2024. \texttt{doi:10.13139/OLCF/2441156}.
\bibitem{croot} X.\ Croot, K.\ Nowrouzi, C.\ Spitzer, C.\ G.\ Almudever, A.\ Blais,
M.\ Carroll, \emph{et al.}, ``Enabling Technologies for Scalable Superconducting Quantum
Computing,'' \emph{arXiv:2512.15001}, 2025.
\bibitem{kennedy} O.\ W.\ Kennedy, W.\ Ahmad, R.\ Armstrong, A.\ Awawdeh, A.\ Bose,
K.\ G.\ Crawford, \emph{et al.}, ``Design and Operation of Wafer-Scale Packages Containing
$>$500 Superconducting Qubits,'' \emph{arXiv:2602.12773}, 2026.
\bibitem{kawabata} S.\ Kawabata, ``Integration and Resource Estimation of Cryoelectronics
for Superconducting Fault-Tolerant Quantum Computers,'' \emph{arXiv:2601.03922}, 2026.
\bibitem{liuphotonic} B.\ Liu and Z.\ R.\ Huang, ``A Cryogenic Hybrid Photonic/CMOS
Controller Architecture for Scalable Superconducting Qubit Control,''
\emph{arXiv:2606.10114}, 2026.
\bibitem{hrl} HRL Quantum Team \emph{et al.}, ``A digitally controlled silicon quantum
processing unit,'' \emph{arXiv:2604.16216}, 2026.
\bibitem{rubin} A.\ H.\ Rubin, V.\ A.\ Norman, and M.\ Radulaski, ``Cryogenic Systems for
Quantum Photonic Technologies: A Practical Review,'' \emph{arXiv:2605.12285}, 2026.
\bibitem{bracevib} N.\ Brace, A.\ D'Addabbo, S.\ D'Eramo, S.\ H.\ Fu, M.\ T.\ Hurst,
T.\ O'Donnell, \emph{et al.}, ``Vibrational sensing at mK temperatures in dry dilution
refrigerators using commercial accelerometers for diverse fundamental physics
applications,'' \emph{arXiv:2601.08817}, 2026.
\bibitem{xieauto} Y.\ Xie, K.\ He, and A.\ Castellanos-Gomez, ``Toward Full Autonomous
Laboratory Instrumentation Control with Large Language Models,'' \emph{arXiv:2604.03286},
2026.
\bibitem{lopezpaz2017} D.\ Lopez-Paz and M.\ Oquab, ``Revisiting Classifier Two-Sample
Tests,'' \emph{International Conference on Learning Representations (ICLR)}, 2017.
\bibitem{breiman2001} L.\ Breiman, ``Random Forests,'' \emph{Machine Learning}, vol.\ 45,
no.\ 1, pp.\ 5--32, 2001.
\bibitem{lightgbm} G.\ Ke, Q.\ Meng, T.\ Finley, T.\ Wang, W.\ Chen, W.\ Ma, Q.\ Ye, and
T.-Y.\ Liu, ``LightGBM: A Highly Efficient Gradient Boosting Decision Tree,'' in
\emph{Advances in Neural Information Processing Systems (NeurIPS)}, vol.\ 30, 2017.
\bibitem{timesnet} H.\ Wu, T.\ Hu, Y.\ Liu, H.\ Zhou, J.\ Wang, and M.\ Long, ``TimesNet:
Temporal 2D-Variation Modeling for General Time Series Analysis,''
\emph{arXiv:2210.02186}, ICLR 2023.
\bibitem{anomalytransformer} J.\ Xu, H.\ Wu, J.\ Wang, and M.\ Long, ``Anomaly Transformer:
Time Series Anomaly Detection with Association Discrepancy,'' \emph{arXiv:2110.02642},
ICLR 2022.
\bibitem{clopperpearson1934} C.\ J.\ Clopper and E.\ S.\ Pearson, ``The Use of Confidence
or Fiducial Limits Illustrated in the Case of the Binomial,'' \emph{Biometrika}, vol.\ 26,
no.\ 4, pp.\ 404--413, 1934.
\bibitem{mcnemar1947} Q.\ McNemar, ``Note on the Sampling Error of the Difference Between
Correlated Proportions or Percentages,'' \emph{Psychometrika}, vol.\ 12, no.\ 2,
pp.\ 153--157, 1947.
\end{thebibliography}
\end{document}